\documentclass{article}

     \PassOptionsToPackage{sort, numbers}{natbib}

\usepackage[preprint]{neurips_2023}

\usepackage[utf8]{inputenc} 
\usepackage[T1]{fontenc}    
\usepackage{hyperref}       
\usepackage{url}            
\usepackage{booktabs}       
\usepackage{amsfonts}       
\usepackage{nicefrac}       
\usepackage{microtype}      
\usepackage{xcolor}         

\belowdisplayskip \abovedisplayskip

\newlength{\sectionReduceTop}
\newlength{\sectionReduceBot}
\newlength{\subsectionReduceTop}
\newlength{\subsectionReduceBot}
\newlength{\abstractReduceTop}
\newlength{\abstractReduceBot}
\newlength{\captionReduceTop}
\newlength{\captionReduceBot}
\newlength{\subsubsectionReduceTop}
\newlength{\subsubsectionReduceBot}

\newlength{\eqnReduceTop}
\newlength{\eqnReduceBot}

\newlength{\horSkip}
\newlength{\verSkip}

\newlength{\figureHeight}
\usepackage{graphicx} 
\usepackage{subcaption}
\usepackage{multicol}
\usepackage{booktabs} 
\usepackage{algorithm} 
\usepackage{algorithmic} 
\usepackage{color}
\usepackage{amsmath}
\usepackage{amssymb}
\usepackage{amsfonts}
\usepackage{multirow, varwidth}
\usepackage{stfloats} 
\usepackage{verbatim}
\makeatletter
\newif\if@restonecol
\makeatother
\usepackage{xr}
\usepackage[amsmath,thmmarks]{ntheorem}
\usepackage{wrapfig}
\usepackage{enumitem}

\usepackage{times}
\usepackage{latexsym}
\usepackage[T1]{fontenc}
\usepackage[utf8]{inputenc}
\usepackage{microtype}
\usepackage{inconsolata}

\usepackage{url}

\usepackage{xspace}
\makeatletter
\DeclareRobustCommand\onedot{\futurelet\@let@token\@onedot}
\def\@onedot{\ifx\@let@token.\else.\null\fi\xspace}

\def\eg{\emph{e.g}\onedot} 
\def\ie{\emph{i.e}\onedot}

\def\etc{\emph{etc}\onedot}

\makeatother

\newcommand{\emptydraft}[1]{{\color{black}{}}}

\newcommand{\abr}[1]{\textsc{#1}}

\usepackage{booktabs,arydshln}
\makeatletter
\def\adl@drawiv#1#2#3{%
        \hskip.5\tabcolsep
        \xleaders#3{#2.5\@tempdimb #1{1}#2.5\@tempdimb}%
                #2\z@ plus1fil minus1fil\relax
        \hskip.5\tabcolsep}
\newcommand{\cdashlinelr}[1]{%
  \noalign{\vskip\aboverulesep
           \global\let\@dashdrawstore\adl@draw
           \global\let\adl@draw\adl@drawiv}
  \cdashline{#1}
  \noalign{\global\let\adl@draw\@dashdrawstore
           \vskip\belowrulesep}}
\makeatother

\usepackage[misc]{ifsym} 
\newcommand*\samethanks[1][\value{footnote}]{\footnotemark[#1]}

\newcommand{\mnf}{{Any-Modality Augmented Language Model}\xspace} 
\newcommand{\mn}{AnyMAL\xspace} 

\newcommand{\dnf}{Multimodal Instruction Tuning} 
\newcommand{\dn}{\abr{MM-IT}\xspace} 

\title{{\mn}: An Efficient and Scalable Any-Modality Augmented Language Model}

\author{
Seungwhan Moon\thanks{Joint First Authors. \hspace{1pt} \Letter: \texttt{\{shanemoon,andreamad8,zhaojiang,tusharn\}@meta.com}} \quad Andrea Madotto\samethanks \quad Zhaojiang Lin\samethanks \quad Tushar Nagarajan\samethanks \\
\textbf{Matt Smith} \quad \textbf{Shashank Jain} \quad \textbf{Chun-Fu Yeh} \quad \textbf{Prakash Murugesan} \\
\textbf{Peyman Heidari} \quad \textbf{Yue Liu} \quad \textbf{Kavya Srinet} \quad \textbf{Babak Damavandi} \quad \textbf{Anuj Kumar} \\
FAIR, Meta \& Meta Reality Labs
}
\date{}

\begin{document}

\maketitle

\vspace{-10pt}
\begin{abstract}

We present \mnf (\mn), a unified model that reasons over diverse input modality signals (\ie text, image, video, audio, IMU motion sensor), and generates textual responses.
%
\mn inherits the powerful text-based reasoning abilities of the state-of-the-art LLMs including LLaMA-2 (70B), and converts modality-specific signals to the joint textual space through a pre-trained aligner module.
%
%
To further strengthen the multimodal LLM's capabilities, we fine-tune the model with a multimodal instruction set manually collected to cover diverse topics and tasks beyond simple QAs.
%
%
We conduct comprehensive empirical analysis comprising both human and automatic evaluations, and demonstrate state-of-the-art performance on various multimodal tasks.


\end{abstract}

\section{Introduction}
\label{sec:introduction}


Large Language Models (LLMs), known for their substantial size and complexity, have significantly enhanced the capacity of machines to understand and articulate human language.
The progress in LLMs has also led to notable advancements in the vision-language domain ~\cite{tsimpoukelli2021multimodal,flamingo, blip2, OpenAI2023GPT4TR}, bridging the gap between image encoders and LLMs to combine their reasoning capabilities.
Prior multimodal LLM research has concentrated on models that combine text and one other modality~\cite{blip2, idefics}, such as text and image models, or has centered on proprietary language models that are not open sourced~\cite{flamingo, OpenAI2023GPT4TR}.

\begin{figure*}[t]
    \centering
    \includegraphics[width=1.00\columnwidth]{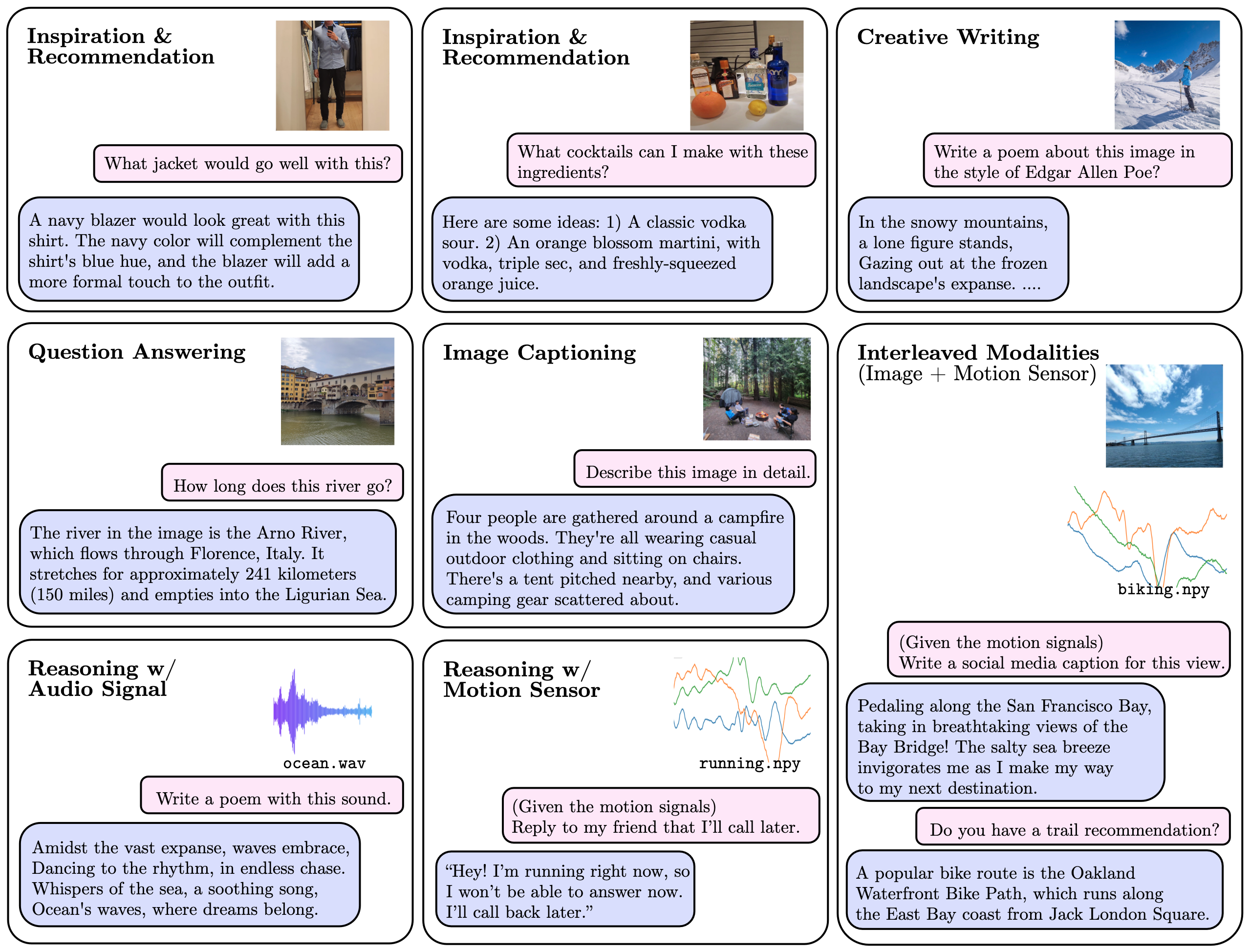}
    \caption{Example \mn outputs. The model understands various input signals (\ie vision, audio, motion sensor signals), and responds to free-form user queries. When multiple modalities are interleaved and given as input (\eg right-most: image + IMU motion sensor signals), the model reasons over them jointly.}
    \label{fig:teaser}
\end{figure*}

To tackle the previously mentioned challenges, we introduce \textbf{\mnf} (\mn) — a collection of multi-modal encoders trained to transform data from various modalities, including images, videos, audio, and IMU motion sensor data, into the text embedding space of an LLM. To achieve this, we extend the work by \cite{tsimpoukelli2021multimodal} to (1) more capable instruction-tuned LLMs (\ie LLaMA-2-70B-chat \cite{llama2}), (2) larger pre-trained modality encoders, and (3) advanced projection layers to handle variable input lengths. 
The model output examples are shown in Figure \ref{fig:teaser}, and an illustration of the overall methodology is shown in Figure \ref{fig:diagram}.  

The key contributions of the work are as follows:
\begin{itemize}
    \item We present an efficient and scalable solution for building Multimodal LLMs. We provide projection layers pre-trained on large datasets with diverse modalities (\eg 200$M$ images, 2.2$M$ audio, 500$K$ IMU time-series, 28$M$ videos) all aligned to the same LLM (LLaMA-2-70B-chat), thus enabling interleaved multimodal in-context prompting.
    \item We further fine-tune the model with the multimodal instruction set across three modalities (image, video, and audio) covering diverse unconstrained tasks beyond simple QA domains. The dataset features high-quality manually collected instruction data, which we thus also use as a benchmark for complex multimodal reasoning tasks.
    \item Our best model achieves strong zero-shot performance in both automatic and human evaluation on diverse tasks and modalities, setting new SOTA with \underline{+7.0\%} relative accuracy improvement on VQAv2, \underline{+8.4\%} CIDEr on zeroshot COCO image captioning, and \underline{+14.5\%} CIDEr on AudioCaps, when compared with the models available in the literature. 
\end{itemize}

%

\section{Related Work}
\label{sec:related_work}

\noindent{\textbf{Large Language Models (LLM)}}:
There has been a surge of LLMs with varying model sizes recently, showcasing remarkable reasoning capabilities.
While the most well-known commercial service is ChatGPT \cite{GPT2, OpenAI2023GPT4TR}, the open-sourced models include FlanT5 \cite{flant5}, GPT-J \cite{gpt-j}, OPT \cite{opt}, LLaMA \cite{llama}, Vicuna \cite{vicuna}, and more recently, LLaMA-2 \cite{llama2}.

Our work builds upon the powerful text-based reasoning capabilities of these LLMs, extending these capabilities to multimodal inputs.

\noindent{\textbf{Vision-Language Models}}:
Numerous studies have addressed the task of instructing a unified model that integrates both visual and linguistic elements, finding practical implementations in domains like image captioning \cite{xu2015show} and visual question answering (VQA) tasks \cite{vqa, visdial, vision-language-navigation}.
While the relative scarcity of data sources aligning different modalities has conventionally been considered the bottleneck in scaling, recent works have shifted towards harnessing the capabilities of pre-trained LLMs, tapping into the knowledge accrued from extensive textual corpora.
These work include Flamingo \cite{flamingo}, OpenFlamingo \cite{openflamingo}, Palm-E \cite{palme}, BLIP-2 \cite{blip2}, InstructBLIP \cite{instructblip}, LLaVA \cite{llava}, IDEFICS \cite{idefics}, MiniGPT-4 \cite{minigpt4} and many more \cite{otter,mplug,multimodalGPT,lamaadapterv2, zhang2023video, su2023pandagpt,lyu2023macaw}, where each model uses different variants of base LLMs.
These models typically undergo fine-tuning stages as well, re-purposing several task-specific vision-language datasets \cite{llava, m3it}.

Our work extends the previous approaches by (1) allowing for diverse input modalities beyond vision signals, (2) presenting a fine-tuning process with our manually collected multimodal instruction tuning data, and (3) scaling the LLM parameters to 70$B$ via an efficient pre-training approach.

\section{Methods}
\label{sec:methods}

\subsection{Pre-training}
\label{subsec:methods:pretraining}

\begin{figure}[t]
    \centering
    \includegraphics[width=1\columnwidth]{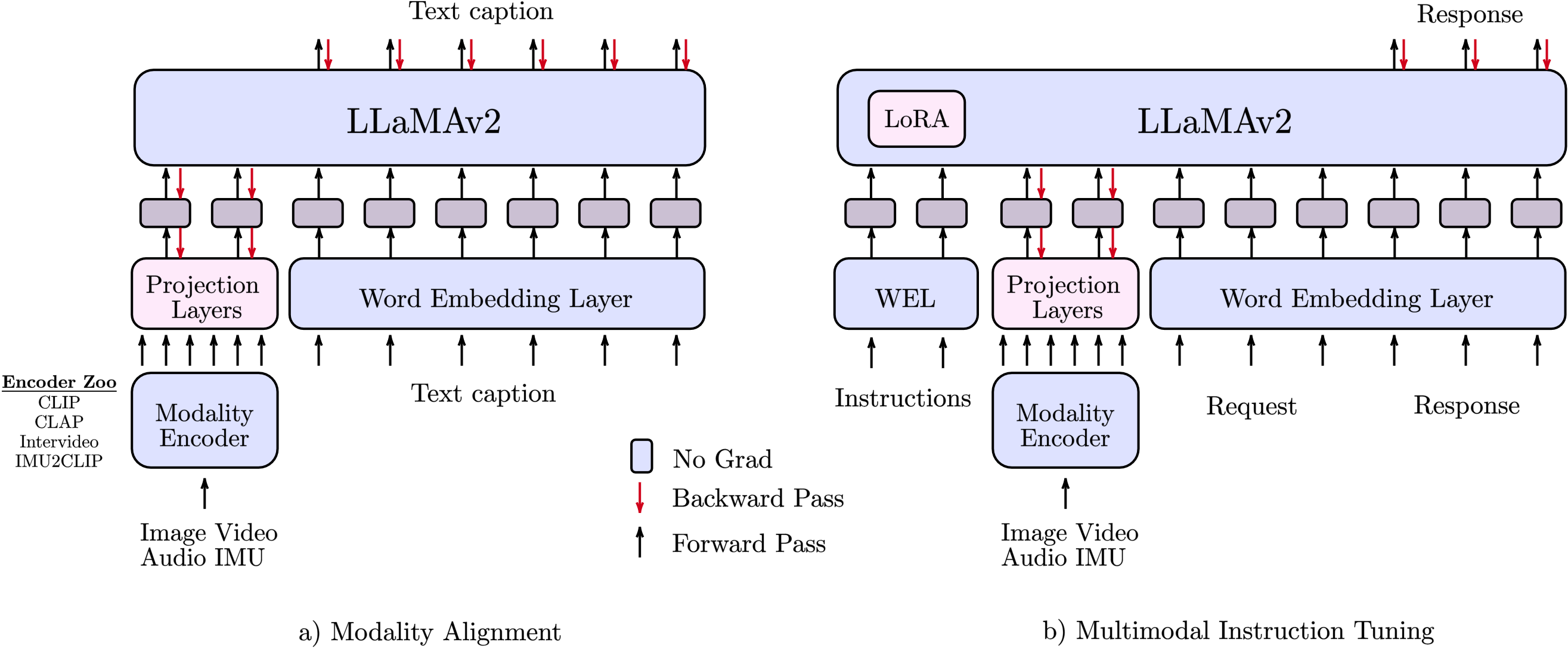}
    \caption{
    \mn Training. (a) Modality alignment pre-training allows for mapping the output of each modality encoder into the joint LLM embeddings space through projection layers. (b) With multimodal instruction tuning, the model learns to associate system instructions and text queries with input multimodal contexts.
    Our modality-specific encoder zoo includes: CLIP ViT-L, ViT-G, DinoV2 (image), CLAP (audio), IMU2CLIP (IMU motion sensor), and Intervideo (video).
    }
    \label{fig:diagram}
\end{figure}

\noindent \textbf{Modality Alignment}: We achieve the multimodal understanding capabilities by pre-training LLMs with paired multimodal data (modality-specific signals and text narrations) (Figure \ref{fig:diagram}).
Specifically, we train a lightweight adapter for each modality to project the input signals into the text token embedding space of a specific LLM.
In this way, the text token embedding space of the LLM becomes a joint token embedding space, with tokens representing either text or other modalities.
The number of token embeddings used to represent each input modality is fixed per adapter, ranging from 64 - 256 in this work.
During the alignment training, we freeze the model parameters of the underlying LLM, which allows it to reach convergence faster than training end-to-end from scratch, and to inherit the reasoning capabilities of the LLM at inference time.
In addition, to maximize the feature compatibility, for each modality we use an encoder $g(\cdot)$ that has already been aligned to a text embeddings space, \eg CLIP \cite{clip, laion} for images, CLAP \cite{clap} for Audio signals, or IMU2CLIP \cite{imu2clip} for IMU signals. 
For each text caption and modality pair $(\mathbf{X}_{\texttt{text}}, \mathbf{X}_{\texttt{modality}})$, we align them using the following objectives with a projection module (\ie Perceiver Resampler \cite{flamingo} for vision encoder, and linear layers for other modalities).
\vspace{-5pt}
\begin{align}
    p(\mathbf{X}_{\texttt{text}} | \mathbf{X}_{\texttt{modality}}) = \prod_{i=1}^{L} p_{\theta}(\mathbf{X}^{[i]}_{\texttt{text}} | \mathbf{Z}_{\texttt{modality}}, \mathbf{Z}^{[1:i\small{-}1]}_{\texttt{text}}) \\
    \mathbf{Z}_{\texttt{modality}} = \texttt{Projection}_{\theta}(h_{\texttt{latents}}, g(\mathbf{X}_{\texttt{modality}}))
\end{align}
%


\noindent \textbf{Datasets}:
For image alignment, we use a cleaned subset of the LAION-2B dataset, filtered using the CAT method and with any detectable faces blurred \cite{laionfiltering}.
For audio alignment, we use AudioSet \cite{audioset} (2.1M samples), AudioCaps \cite{kim2019audiocaps} (46K samples), and CLOTHO \cite{clotho} (5K samples) datasets.
We use the Ego4D dataset \cite{ego4d} for IMU and text alignment (528K).

\begin{table*}[!ht]
\centering
\scalebox{0.95}{
    \begin{tabular}{p{4.4cm}p{9.3cm}}
    \toprule
    \multicolumn{2}{l}{\textbf{Example 1. Creative Instructions (Image)}} \\
    \midrule
    \multirow{4}{*}{        
        \begin{minipage}{.6\textwidth}
            \includegraphics[width=4cm]{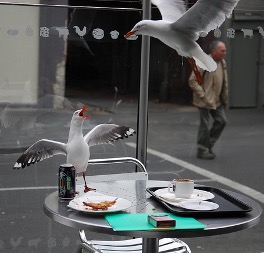}
        \end{minipage}
    }
    &
    \textbf{\underline{Instruction Prompt}}
    \\
    & Write a short story about the scene in this photo. Include some dialogue between the characters.	\\
    & \textbf{\underline{Response}} \\
    & The restaurant table stood empty but adorned with remnants of a meal. A seagull landed on its edge, spreading his wings in a display of dominance. With a raucous cry, he fiercely guarded the plate before him, which he intended to make his dinner. His piercing gaze fixed upon another seagull, wings poised for landing, his eyes locked on the feast. "Back off!" the first seagull screeched. Beak open wide, he emitted a symphony of squawks, challenging the attacker. Undeterred, the second seagull hesitated, torn between his desire for snacks, and fear of the defending seagull. The tension hung in the air, and as the first seagull prepared to yell again, the second replied with a sullen "Fine," and flew off. Triumphant, the first gull tucked into his dinner. \\
    
    \toprule
    \multicolumn{2}{l}{\textbf{Example 2. Detailed Description (Image)}} \\
    \midrule
    
    \multirow{4}{*}{        
        \begin{minipage}{.6\textwidth}
            \includegraphics[width=4cm]{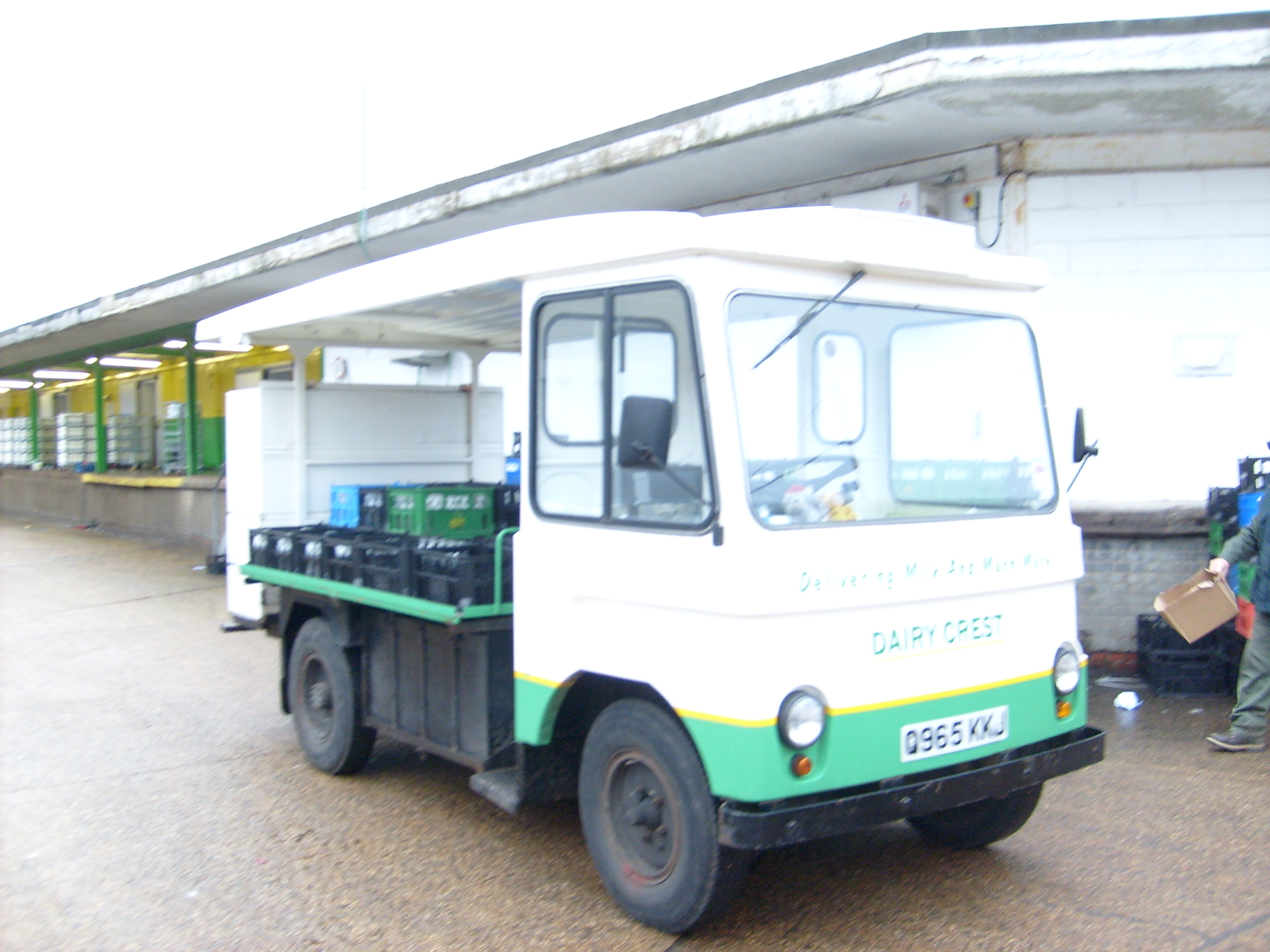}
        \end{minipage}
    }
    &
    \textbf{\underline{Instruction Prompt}}
    \\
    & Explain all the details of the scene, starting from left to right.	\\
    & \textbf{\underline{Response}} \\
    & This is a photo of a small delivery truck. In the background of the photo is a loading dock of some type with green and yellow structures under an awning. The small truck stands at the center of the photo. It has a flatbed and awning over the flatbed and some type of white cabinet on the back and a black storage compartment under. The bed is green and has black and green milk crates stacked on it. The cab of the truck is small and tall with large windows all around. It is white with a green lower section and fine yellow stripe separating the green and white paint. The front of the truck has `Dairy Crest' painted on it in green and has a European style license plate. There is a partial image of a man in the background. \\

    \toprule
    \multicolumn{2}{l}{\textbf{Example 3. Open-ended Question and Answering (Video)}} \\
    \midrule
    
    \multirow{6}{*}{        
        \begin{minipage}{.6\textwidth}
            \includegraphics[width=4cm]{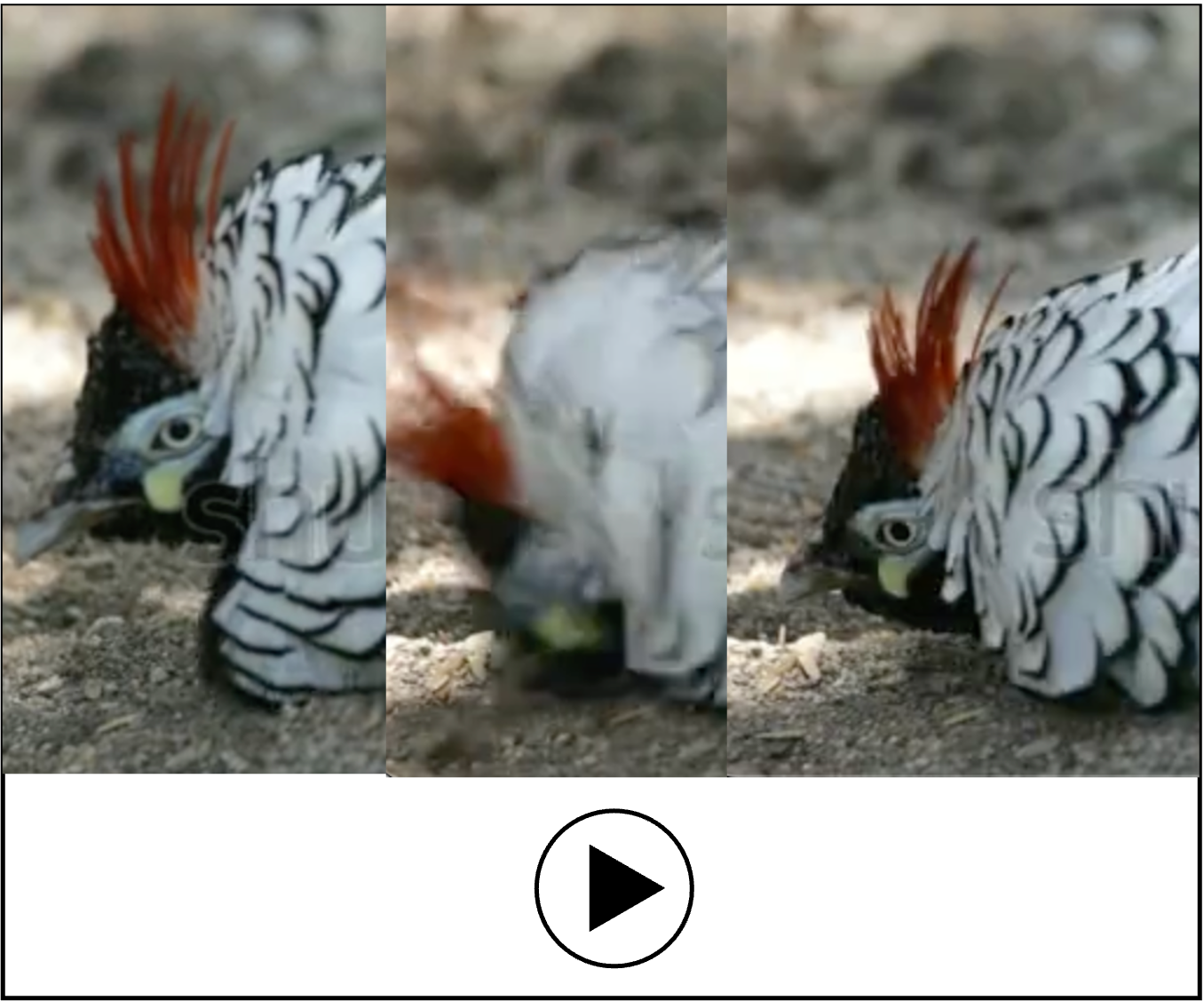}
        \end{minipage}
    }
    &
    \textbf{\underline{Instruction Prompt}}
    \\
    & What is this animal doing and why are they doing it?\\
    & \textbf{\underline{Response}} \\
    & Pheasants eat gravel to help them in the digestion of their food. This helps them survive multiple weeks without feeding during extreme weather. A newly hatched chick will survive on insects for a few weeks of its life. \\
    & \\
    & \\

    \toprule
    \multicolumn{2}{l}{\textbf{Example 4. Multimodal Reasoning (Audio)}} \\
    \midrule    

    \multirow{4}{*}{        
        Audio File (\href{https://freesound.org/people/Timbre/sounds/94571/}{Freesound Link})
    }
    &
    \textbf{\underline{Instruction Prompt}}
    \\
    & What could be the likely momentous occasion in this sound clip?	\\
    & \textbf{\underline{Response}} \\
    & This is a countdown to a rocket taking off into space by a robotic voice.\\ 
    \bottomrule
    \end{tabular}
}
\caption{
Example data from our \dn dataset, manually collected instruction prompt and response pairs for diverse modalities (\ie image, video, audio). The collected instructions cover diverse domains (\eg creative writing, open-ended reasoning), and are strictly grounded to the provided multimodal context (\ie queries \textit{cannot} be answered without understanding the accompanying multimodal context).
The \dn data serves as both a fine-tuning dataset as well as an evaluation benchmark for complex multimodal reasoning tasks.
}
\label{tab:mit_examples}
\end{table*}

\noindent \textbf{Quantization}:
Scaling the pre-training to 70B parameter models for a large dataset (200M+ instances) requires significant resources, often requiring an FSDP \cite{zhao2023pytorch} wrapper to shard the model across multiple GPUs.
To effectively scale our training, we implement the quantization strategies (4 bits and 8 bits) \cite{qlora} in our multimodal settings, in which we keep the LLM component of our model frozen and only have the modality tokenizers trainable.
This approach shrinks the memory requirement by an order of magnitude.
Thus, we are able to train 70B \mn on a single 80GB VRAM GPU with a batch size of 4.

Compared with FSDP, we observe that the proposed quantization approach achieves the same throughput while only using half of the GPU resources.
We do note that the training / validation loss were constantly higher compared to the FSDP training, but nevertheless did not impact the generation quality (at inference time, we use the original LLM at full precision to maximize the accuracy).

\subsection{Fine-tuning with Multimodal Instruction Datasets}
\label{subsec:methods:finetuning}

To further improve the model's instruction-following capability with respect to diverse input modalities, we perform additional fine-tuning with our multimodal instruction-tuning (\dn) dataset.
Specifically, we concatenate the input as $\texttt{[<instruction> <modality\_tokens>]}$, such that the response target is grounded on both textual instructions and the modality input.
%
%
We perform ablations over (1) training the projection layers without altering the LLM parameters, or (2) using Low-Rank Adaptation \cite{lora} to further tune the LM behaviors.

We use both manually collected instruction-tuning datasets and synthetic data.

\noindent \textbf{Manual Annotation.}
While there are publicly available third-party datasets on various VQA tasks, we observe that many of these data have insufficient diversity and quality — in particular for aligning LLMs towards diverse multimodal instruction-following tasks that go much beyond simple QA queries (\eg ``\textit{Create a poem using this image}'', ``\textit{Extract the phone number on this flyer}''). 


Therefore, we focus on collecting 60K examples of high-quality multimodal instruction tuning data for multiple modalities, as illustrated in Table \ref{tab:mit_examples}.
Specifically, we use various Creative Commons licensed, publicly available images, 
and augment these images with manually created instructions and responses.
Annotators are required to provide instruction and answer pairs that are strictly multimodal, such that queries cannot be answered without understanding the accompanying multimodal context.

We show that our results notably improve using these fewer but well-balanced and higher-quality examples from our own vendor-based annotation efforts.

\noindent \textbf{Synthetic Augmentation.}
In addition to the high-quality ground-truth instruction tuning data above, we augment the dataset using the LLaMA-2 (70$B$) \cite{llama2} model, following similar approaches proposed by LLaVA \cite{llava}.
Specifically, we use a textual representation of the image (\ie multiple captions, bounding boxes information and objects) to generate question-answer pairs for the image. 
We generate 150K image-instruction-response pairs on varying domains and question types. 

Note that our process strictly uses only open-sourced models 
-- as opposed to other works that use commercial services such as ChatGPT or GPT-4.


    

\section{Experiments}
\label{sec:experiments}

\subsection{Tasks}
\label{subsec:experiments:tasks}

We evaluate the model's performance on two categories of tasks in the zero-shot setting: (1) captioning tasks for various modalities, and (2) multimodal reasoning and instruction-following tasks.

\noindent \textbf{Captioning Tasks}.
We evaluate \mn's primary capability of generating captions given input modalities, which is aligned with the pre-training objective.
The main purpose of the captioning task is to understand the alignment level between the text and other modalities after pre-training.
Since the captioning tasks typically don't require secondary reasoning steps, we expect that LLM weights or parameter sizes have less influence on the task. 

\noindent \textbf{Multimodal Reasoning Tasks}.
Given the high-level of alignment among the modalities, we evaluate the model's reasoning and instruction-following abilities which it inherits from the core instruction-tuned LLM, as well as from the multimodal instruction-tuning process.

We conduct a comprehensive comparison with strong baseline models for each respective modality pair (vision-language and audio-language) from the open-sourced literature.

\textit{Note}: As the \dn datasets include some in-domain images from public benchmarks (\eg COCO), we report results separately for the pre-trained models (without further instruction tuning in Section \ref{subsec:methods:finetuning}) and the instruction-tuned models -- to denote a \textit{strict} zeroshot setup.
All multimodal-instruction-tuned \mn models are marked with ``\dn{}'' in the following sections.


\begin{table}[t]

    \centering
    \scalebox{0.99}{
        \begin{tabular}{@{}ccc@{}}
        \toprule
        \multirow{2}{*}{\textbf{Models}} &  \multicolumn{1}{c}{\textbf{COCO}} &  \multicolumn{1}{c}{\textbf{\dn-Cap}} \\ 
            \cmidrule(r){2-2}
            \cmidrule(r){3-3}
          & CIDEr & CIDEr \\  
        \midrule
        \midrule
        \multicolumn{1}{l}{BLIP-2} \cite{blip2}   & - & 2.9 \\
        \multicolumn{1}{l}{MiniGPT4} \cite{minigpt4}    & - & 14.1 \\
        \multicolumn{1}{l}{LLaVA} \cite{llava}   & - & 14.3 \\        
        \multicolumn{1}{l}{CM3Leon} \cite{cm3leon}    & 61.6 & - \\
        \multicolumn{1}{l}{OpenFlamingo-v2 9B}  \cite{openflamingo}    & 79.5 &  1.8 \\
        \multicolumn{1}{l}{Flamingo-3B}  \cite{flamingo}    & 73.0 & -  \\
        \multicolumn{1}{l}{Flamingo-9B}  \cite{flamingo}   & 79.4 &  -  \\
        \multicolumn{1}{l}{Flamingo-80B} \cite{flamingo}    & 84.3 & -  \\
        \multicolumn{1}{l}{IDEFICS-9B} \cite{idefics}   & 46.0 & - \\
        \multicolumn{1}{l}{IDEFICS-80B} \cite{idefics}    & 91.8 & - \\        
        \midrule
         \multicolumn{1}{l}{\mn 13B (ViT-G)}  & \textbf{99.5} &  \underline{15.5} \\
         \multicolumn{1}{l}{\mn 70B (ViT-G)}   & \underline{95.9} & \textbf{15.7} \\
        \bottomrule
        \end{tabular}
    }
     \caption{
            \textbf{Zeroshot \underline{Image} Captioning} performance on COCO and \dn-Cap.
            Ablations (bottom) over our \mn with varying LLM sizes.
            \textbf{Bold} and \underline{underlined} denote the top and the second-best performance, respectively.
            ``-'': the model (a) does not report results on the marked benchmarks, or (b) is pretrained or fine-tuned on the respective dataset, thus not suitable for the zeroshot evaluation above.
            \mn demonstrates the state-of-the-art zeroshot visual understanding capabilities compared to the baseline vision-language models.
        }
    \label{tab:exp:image_captioning}    
\end{table}

\begin{figure}[t]
    \centering
    \includegraphics[width=1.0\columnwidth]{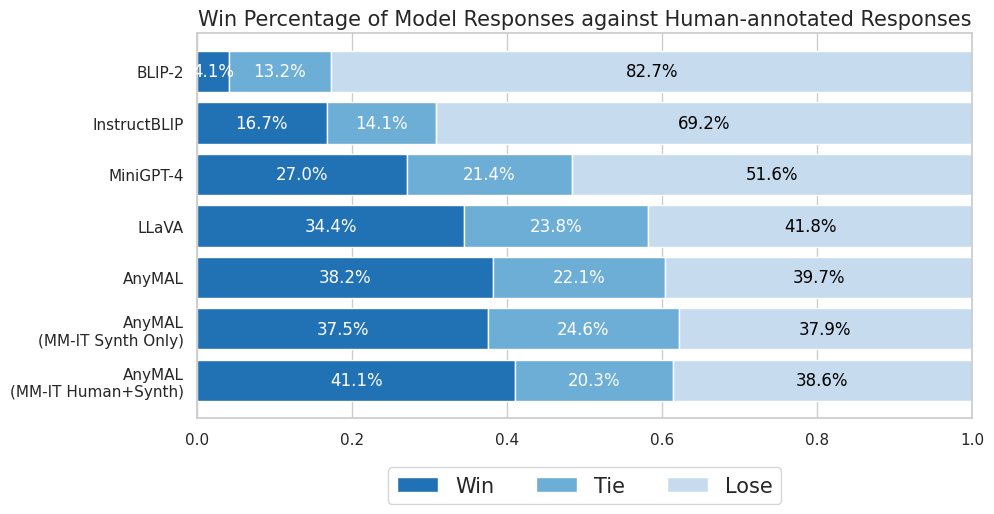}
    \caption{\textbf{\underline{Image}-based reasoning} human evaluation results on pairwise comparisons (\% win, tie and lose) with baseline outputs \textit{against} the manually annotated ground-truth samples from \dn (1K test set).
    Baselines used: BLIP-2 (FlanT5$_{\texttt{XXL}}$) \cite{blip2}, InstructBLIP (Vicuna-13B) \cite{instructblip}, MiniGPT4 \cite{minigpt4} and LLaVA \cite{llava}.
    \mn demonstrates a smaller gap with human-generated responses (41.1\% win), compared to the baselines (LLaVA: 34.4\% win, and MiniGPT4: 27.0\%). 
    }
    \label{fig:human_eval}
\end{figure}

\subsection{Quantitative Analysis}
\label{subsec:experiments:quantitative}

\begin{table}[t]
\centering
\scalebox{0.95}{
    \begin{tabular}{@{}lccc@{}}
    \toprule
    \textbf{Models} &\textbf{Response Acc} & \textbf{Obj Recognition} & \textbf{Integrity} \\
    \midrule
    \midrule
    \multicolumn{1}{l}{BLIP-2 (FlanT5$_{\texttt{XXL}}$)} \cite{blip2} & 43.3 & 73.5 & 99.3 \\
    \multicolumn{1}{l}{InstructBLIP (Vicuna-13B)} \cite{instructblip} & 46.3 & 73.2 & 98.3 \\
    \multicolumn{1}{l}{Mini-GPT4} \cite{minigpt4} & 42.7 & 73.0 & \underline{99.5} \\
    \multicolumn{1}{l}{LLaVA} \cite{llava} & 51.7 & \textbf{85.4} & \underline{99.5} \\
    \midrule
    \mn 70B                    & \underline{56.0} & 82.4 & 99.3 \\
    \mn 70B (\dn Synth Only)        & 54.2 & \underline{83.5} & \underline{99.5} \\
    \mn 70B (\dn Human+Synth)  & \textbf{58.0} & 79.3 & \textbf{99.7} \\
    \bottomrule
    \end{tabular}
}
\caption{
    \textbf{\underline{Image}-based Reasoning} human evaluation results on 1K test set from \dn on different axes: (a) Response Accuracy and Relevance (\%) -- whether responses are relevant to instructions and factually correct without any hallucinations, (b) Object Recognition (\%) -- whether key objects are identified at a detailed level, and (c) Integrity (\%) -- whether responses include offensive language.
    \dn indicates the model that has been instruction-tuned either with synthetic data only, or with the manually collected set (Section \ref{subsec:methods:finetuning}).
}
\label{tab:human_eval_likert}
\end{table}

\begin{table}[t]

    \centering
    \scalebox{0.89}{
        \begin{tabular}{@{}ccccccc@{}}
        \toprule
        \multirow{2}{*}{\textbf{Models}} &  \multicolumn{1}{c}{\textbf{H-Meme}} &  \multicolumn{1}{c}{\textbf{VQAv2}} & 
        \multicolumn{1}{c}{\textbf{TextVQA}} & \multicolumn{1}{c}{\textbf{S-QA}}  & \multicolumn{1}{c}{\textbf{VizWiz}}  & \multicolumn{1}{c}{\textbf{OKVQA}}\\ 
            \cmidrule(r){2-2}
            \cmidrule(r){3-3}
            \cmidrule(r){4-4}
            \cmidrule(r){5-5}
            \cmidrule(r){6-6}
            \cmidrule(r){7-7}
          & AUC & Accuracy & Accuracy & Accuracy & Accuracy & Accuracy\\  
        \midrule
        \midrule
        \multicolumn{1}{l}{OpenFlamingo-v2} \cite{openflamingo} & 51.6  & 50.5 & 24.2 & - & 27.5 & 37.8 \\
        \multicolumn{1}{l}{Flamingo-3B} \cite{flamingo}  & 53.7 & 49.2 & 30.1 & - & 28.9 & 41.2\\
        \multicolumn{1}{l}{Flamingo-9B} \cite{flamingo}  & 57.0 & 51.8 & 31.8 & - & 28.8 & 44.7\\
        \multicolumn{1}{l}{Flamingo-80B} \cite{flamingo}  & 46.4 & 56.3 & \underline{35.0} & - & 31.6 & \textbf{50.6}\\
        \multicolumn{1}{l}{BLIP-2 (FlanT5$_{\texttt{XXL}}$)} \cite{blip2}  & 52.0 & 65.0$^\dagger$ & 44.1* & 64.5 & 29.4 & 45.9 \\
        \multicolumn{1}{l}{InstructBLIP (V-13B)} \cite{instructblip}  & 53.7 & - & 50.7$^\dagger$* & \underline{70.6} & 33.4 & - \\
        \multicolumn{1}{l}{IBELICS-9B} \cite{idefics} & 51.8  & 50.9 & 25.9 & - & 35.5 & 38.4\\
        \multicolumn{1}{l}{IBELICS-80B} \cite{idefics} & 60.6 & \underline{60.0} &  30.9 & - & \underline{36.0} & 45.2 \\
        \midrule
         \multicolumn{1}{l}{\mn 13B (ViT-G)} & 66.0 & 59.6 & 24.7 & 52.7 & 24.4 &  33.1\\
         \multicolumn{1}{l}{\mn 70B (DINO-V2)} & 65.6 & 59.2 & 13.7 & 64.7 & 23.6 & 41.4 \\
         \multicolumn{1}{l}{\mn 70B (ViT-L)} & \underline{68.2} & 62.0 & \textbf{35.4} & 67.2 & 32.2 & 41.2 \\
         \multicolumn{1}{l}{\mn 70B (ViT-G)} & \textbf{69.1} & \textbf{64.2} & 32.9 & \textbf{70.8} & 33.8 & 42.6 \\
         \midrule
         \multicolumn{1}{l}{\mn 70B (\dn;ViT-G)} & 67.4 & 67.8$^\dagger$ & 32.5 & 67.6 & \textbf{41.3} & \underline{46.1} \\         
        \bottomrule
        \end{tabular}
        }

     \caption{
            \textbf{Zeroshot \underline{Image}-based QA} results on 6 different VQA datasets (H-Meme: Hateful Meme, S-QA: Science QA).
            Ablations (bottom) over \mn with varying base ViTs and LLM sizes.
            \dn (last row) denotes the model fine-tuned on our instruction dataset.
            \textbf{Bold} and \underline{underlined} denote the top and the second-best performance, respectively.
            \mn demonstrates competitive zeroshot multimodal reasoning capabilities, compared to the baseline vision-language models.
            *: Results with additional OCR inputs.
            $\dagger$: in-domain images (\ie COCO, TextCap) have been used during training, thus not a strict zeroshot performance.            
        }
    \label{tab:exp:vqa}    
\end{table}

\begin{table}[t]
\centering
    \scalebox{0.99}{
        \begin{tabular}{ccccc}
        \toprule
        \multirow{2}{*}{\textbf{Models}} &  \multicolumn{3}{c}{\textbf{AudioCaps}} \\ 
            \cmidrule(r){2-4}
          & CIDEr & SPICE & SPICEr \\  
        \midrule
        \midrule
        \multicolumn{1}{l}{TopDown-AlignedAtt} \cite{kim2019audiocaps} & 59.3 &  14.4 & 36.9 \\
        \multicolumn{1}{l}{CNN10-VGG} \cite{xu2021investigating} & 66.0  & 16.8 & 41.4 \\
        \multicolumn{1}{l}{ACT} \cite{mei2021audio} & 67.9 &  16.0 & 42.0 \\
        \multicolumn{1}{l}{PANNs + BERT} \cite{liu2022leveraging} & 66.7 & 17.2 & 42.0 \\
        \midrule
        \multicolumn{1}{l}{\mn 7B (CLAP)} & 70.4  & 21.0 & 45.7 \\
        \multicolumn{1}{l}{\mn 13B (CLAP)} & \underline{72.1} & \underline{22.0} & \underline{47.0} \\
        \multicolumn{1}{l}{\mn 70B (CLAP)} & \textbf{77.8} & \textbf{23.0} & \textbf{50.4} \\
        \bottomrule
        \end{tabular}
    }
\caption{
    \textbf{Zeroshot \underline{Audio} Captioning} results on AudioCaps. Ablations (bottom) over our \mn with varying base LLMs and sizes.
    \mn attains the best performance across multiple metrics, showing the model's strong performance in audio signal understanding.
}
\label{tab:aac}
\end{table}

\begin{table}[t]

    \centering
    \scalebox{0.99}{
        \begin{tabular}{@{}cccc@{}}
        \toprule
        \multirow{2}{*}{\textbf{Models}} &  \multicolumn{1}{c}{\textbf{STAR}} &  \multicolumn{1}{c}{\textbf{How2QA}} & 
        \multicolumn{1}{c}{\textbf{NextQA}}\\ 
            \cmidrule(r){2-2}
            \cmidrule(r){3-3}
            \cmidrule(r){4-4}
          & Accuracy & Accuracy & Accuracy \\  
        \midrule
        \midrule
        \multicolumn{1}{l}{Internvideo (8)} \cite{wang2022internvideo} & 41.6 & 62.2 & 49.1 \\
        \multicolumn{1}{l}{Flamingo-9B} \cite{flamingo}  & 	41.8 & - & - \\
        \multicolumn{1}{l}{Flamingo-80B} \cite{flamingo}  & 39.7 & - & -\\
        \multicolumn{1}{l}{BLIPv2 ViTG FlanT5xxl (4)} \cite{li2023blip} & 42.2  & \textbf{69.8} & \textbf{62.4} \\
        \midrule
         \multicolumn{1}{l}{\mn-Video 13B (Internvideo) (8)} & 37.5 & 54.8 & 46.8 \\
         \multicolumn{1}{l}{\mn-Video 70B (Internvideo) (8)} & 41.3 &  60 & 50.6 \\
         \midrule
         \multicolumn{1}{l}{\mn-Image 13B (ViT-G) (4)} & \underline{44.4} & 59.6 & 47.9 \\
         \multicolumn{1}{l}{\mn-Image 70B (ViT-G) (4)} & \textbf{48.2} & \underline{68.1} & \underline{57.6} \\
        \bottomrule
        \end{tabular}
        }

     \caption{
            \textbf{Zeroshot \underline{Video}-based QA} accuracy on STAR, How2QA, and NextQA. Ablations (bottom) over \mn with image vs video model and LLM sizes.
            \mn demonstrates competitive zeroshot multimodal reasoning capabilities, compared to the baseline vision-language models. Numbers in brackets denote number of frames used per video clip.
        }
    \label{tab:exp:video}    
\end{table}

\noindent \textbf{Image Caption Generation}: 
Table \ref{tab:exp:image_captioning} shows zeroshot image captioning performance on COCO \cite{mscoco} and a subset of the \dn dataset marked with the ``detailed description'' task (\dn-Cap).
It can be seen that our \mn variants significantly outperform the baselines in both datasets.
It is worthwhile to note that there is no significant gap between the performance of the \mn-13B and the \mn-70B variants.
This result indicates that the underlying LLM capability has smaller impact to the image caption generation task (which corresponds to the core visual understanding capability), but is largely dependent on the scale of the data and the alignment methods.
We attribute the slight under-performance of the \mn-70B on COCO to the general verbosity of the LLaMA-70B model, which negatively impacts the score when evaluated against COCO captions that tend to be brief and concise.
As expected, the automatic evaluation on \dn-Cap shows lower CIDEr scores overall, attributed to the much longer response length in detailed descriptions (See Table \ref{tab:mit_examples} for an example).

\noindent \textbf{Human Evaluation on Multimodal Reasoning Tasks}:
\dn features diverse multimodal instruction and ground-truth answer pairs.
We evaluate the performance of our models (pre-trained and instruction-tuned) against other vision-language models publicly available to run and use (\ie LLaVA \cite{llava}, MiniGPT4 \cite{minigpt4}).
Since the responses are subjective in nature (\eg creative writing -- ``\textit{Write a poem about this image}'', we believe that human assessment provides the most precise insight into the performance and capabilities of our proposed model.

We therefore collect pairwise comparisons for each baseline against 1$K$ ground-truth samples (Figure \ref{fig:human_eval}), as well as the Likert scale scores (0-2) for each of the following criteria.
The criteria for preference ranking includes response accuracy, object recognition accuracy, and integrity (see the full rubrics in Appendix \ref{a-sec:human_eval}).
Response accuracy measures whether the response contains the relevant, factually correct and verifiable information (without any hallucinations) with regards to the image and the instruction.
Object recognition accuracy strictly measures whether the key objects are correctly recognized at a detailed level -- primarily concerning the model's visual knowledge.
Finally, the integrity metric measures whether the response shows any harmful or offensive language.

Figure \ref{fig:human_eval} shows that \mn achieves strong performance with a narrower gap against the manually annotated ground-truth samples (41.1\% win), compared to the baselines (LLaVA : 34.4\% win, and MiniGPT4: 27.0\% win). 
Notably, the model fine-tuned with the full instruction set exhibits the highest rate of preferential wins, showing a competitive level of visual understanding and reasoning capabilities comparable to human-annotated responses.
It is also worthwhile to note that BLIP-2 and InstructBLIP suffer on these open-ended queries (4.1\% and 16.7\% preferential win, respectively), despite their strong performance in the public VQA benchmarks (Table \ref{tab:exp:vqa}).

Table \ref{tab:human_eval_likert} shows the detailed breakdown of scores for each criteria.
Specifically, it can be seen that the model instruction-tuned with both manually collected and synthetically curated sets attains the highest response accuracy and relevancy score (12.2\% relative improvement compared to the strongest baseline: LLaVA).
This result highlights the enhanced capability of the model to comprehend and precisely answer questions in accordance with provided instructions.
It is worth mentioning that the model, after instruction tuning, exhibits a decline in its ability to provide detailed recognition and descriptions of objects.
We attribute this result to our intention behind collecting the \mn dataset which promotes concise responses.
Consequently, the model tends to provide more succinct descriptions, rather than elaborating on details (which often increases the risk of factual inaccuracies). 
High integrity scores (99+\%) are achieved by all baselines.

\noindent \textbf{VQA Benchmarks}:
Table \ref{tab:exp:vqa} shows the zeroshot performance on the Hateful Meme dataset \cite{hatefulmeme}, VQAv2 \cite{vqa}, TextVQA \cite{TextVQA}, ScienceQA \cite{scienceqa} (image-paired subset), VizWiz \cite{vizwiz}, and OKVQA \cite{okvqa} compared against the models in the literature that report zeroshot results on the respective benchmark.
We focus on zeroshot evaluation to best estimate the model's performance on the open-ended queries at inference time.

Overall, our \mn exhibits the strongest performance compared to the baselines across multiple tasks.
Notably, the pretrained \mn models show competitive performance even without further fine-tuning on \dn -- demonstrating strong reasoning capabilities retained in the base pre-training phase.
Comparing the two text-guided vision encoders ViT-L \cite{clip} and ViT-G \cite{laion}, ViT-G achieves higher scores in most of the vision datasets. 
These results suggest that larger encoders trained over more steps offer better representations.
However, the TextVQA dataset presents a unique case in which the image encoder's resolution plays a pivotal role in recognizing text within images. Consequently, the ViT-L model, with a resolution of 336x336, achieves a higher score than the ViT-G model, which had a resolution of 224x224.
DinoV2 \cite{dinov2}, which is trained in a discriminative self-supervised approach, shows worse performance compared to the text-guided vision encoders, demonstrating the importance of feature space alignment.
Among the base LLM models, our 70B model demonstrates the most robust performance, underscoring the influence of substantial reasoning proficiency inherent in larger LLMs on tasks involving visual reasoning.


\begin{wrapfigure}{R}{7cm}
    \centering
    \includegraphics[width=0.5\columnwidth]{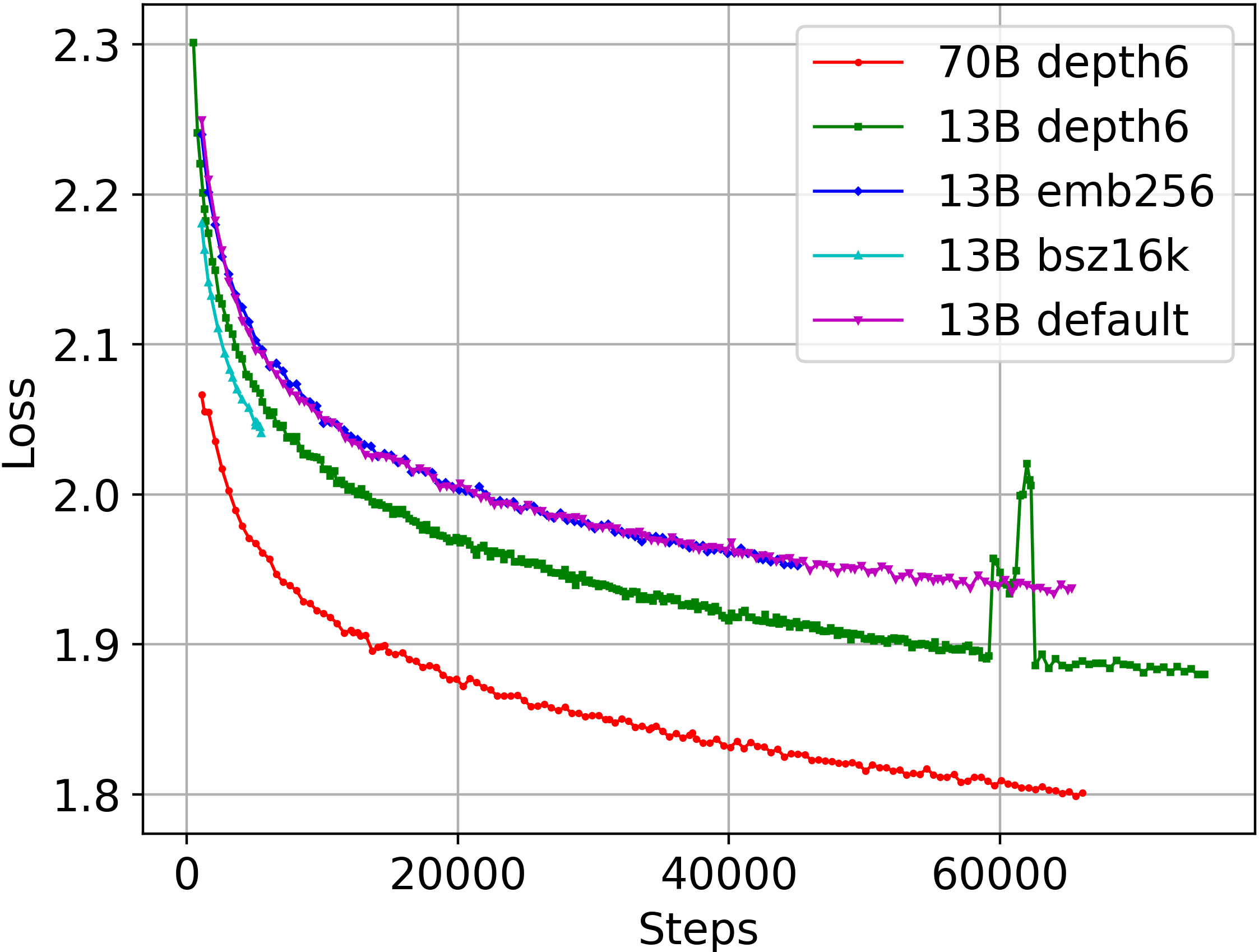}
    \caption{
        Training losses on variants (13B and 70B) of \mn during image-text pre-training.
        Default parameters: batch size: 2,048, resampler layers: 2, the number of modality tokens: 64.
        Each line features an ablation with one differing parameter from the default setting, \ie \texttt{bsz16K}: 16,384 batch size, \texttt{emb256}: 256 visual tokens, \texttt{depth6}: 6 resampler layers. 
    }
    \label{fig:losses}
\end{wrapfigure}

\noindent \textbf{Video QA benchmarks}: We evaluate our model on three challenging video question-answering benchmarks in Table~\ref{tab:exp:video}: How2QA~\cite{li2020hero}, STAR~\cite{wu2021star} and NextQA~\cite{xiao2021next}. Our model demonstrates competitive results compared to the baselines, and achieves state-of-the-art performance on the STAR benchmark. 
Note that we compare against approaches that process the full, untrimmed video clip to generate answers. Prior work has shown additional improvements with careful frame-selection strategies~\cite{yu2023self}. Our approach is compatible with such strategies, however that is beyond the scope of our experiments.
We additionally report model variants trained exclusively on videos from HowTo100M~\cite{miech2019howto100m} coupled with text from ASR transcripts, and using explicit video encoders (Internvideo~\cite{wang2022internvideo}) as opposed to image encoders. However, these models perform worse due to the weak alignment of ASR to video clips and lower diversity in content (i.e., instructional videos). 

\begin{table*}[t]
\centering
\scalebox{0.95}{
    \begin{tabular}{p{4.4cm}p{9.3cm}}
    \toprule

    \textbf{Audio Signals Input} & \textbf{\mn Output} \\
    \toprule

    \multirow{2}{*}{        
        \begin{minipage}{.6\textwidth}
            \includegraphics[width=4cm]{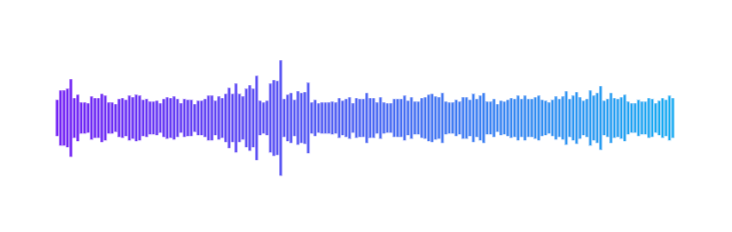}
        \end{minipage}
    }
    & \textbf{Gold Caption}: Typing on a keyboard is occurring in a quiet environment \\
    & \textbf{Prompt}: Can you guess my location? Reply with one or two sentences.  \\
    & \textbf{Response}: Based on the sound of the typing, I would guess that you are located in a quiet, indoor setting, possibly in an office or study. The sound of the typing is clear and crisp, with a slight echo or reverb that suggests a room with a high ceiling or a large, open space. \\
    \midrule

    \multirow{2}{*}{        
        \begin{minipage}{.6\textwidth}
            \includegraphics[width=4cm]{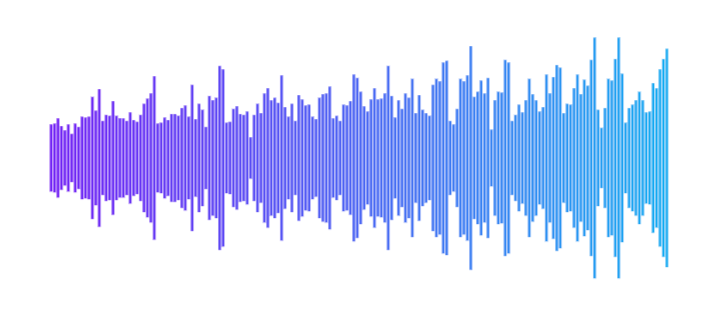}
        \end{minipage}
    }
    & \textbf{Gold Caption}: Very strong wind is blowing, and waves are rustling \\
    & \textbf{Prompt}: Based on the weather, what should I wear? Reply with one or two sentences.  \\
    & \textbf{Response}: Based on the sound of the wind and the waves, it seems like a stormy weather is approaching. It's best to dress in layers and wear something waterproof and warm to stay comfortable and safe. \\

    \midrule

    \multirow{2}{*}{        
        \begin{minipage}{.6\textwidth}
            \includegraphics[width=4cm]{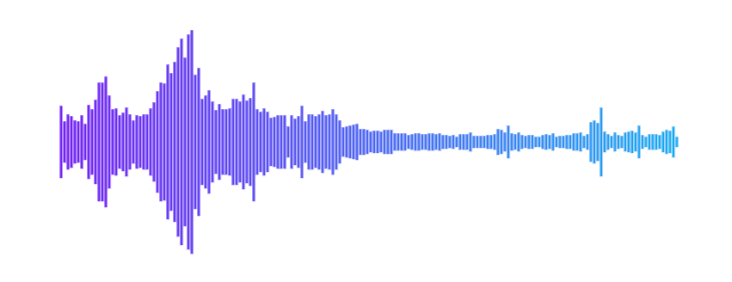}
        \end{minipage}
    }
    & \textbf{Gold Caption}: A baby cries while electronic speech occurs in the background, an adult female speaks, a sneeze occurs, and an adult female speaks once more. \\
    & \textbf{Prompt}: Describe the sound in one or two sentences. \\
    & \textbf{Response}: The sound is a loud, high-pitched scream of a child, with a hint of a whine, as if the child is crying and screaming at the same time. The sound is accompanied by the muffled voices of adults in the background, possibly the child's parents or caregivers, speaking in a soothing tone.\\
    
    \bottomrule
    \end{tabular}
}
\caption{
Example outputs from \mn on unseen \underline{audio} samples. Gold captions are shown as a reference for the audio signals. The text prompt and the audio signals are used as input.
}
\label{tab:audio_examples}
\end{table*}

\noindent \textbf{Ablations on Hyperparameters}: 
Figure \ref{fig:losses} shows the training losses on the variants of \mn during image-text pre-training.
Due to the high computational cost of training 70B models, we conduct ablations only on the 13B models.
We then use the optimal set of hyperparameters from these experiments as the default configuration for training the 70B models.
Since the trainable parameters remain consistent for both the 13B and 70B models, we anticipate that the behavior during model training will remain unchanged.
The following key hyperparameters are considered: the layers of the Resampler  (2 \textit{vs.} 6), the number of multimodal token embeddings to represent visual input (64 \textit{vs.} 256 tokens), and batch size of pre-training (2,048 \textit{vs.} 16,384). Overall, we observed that increasing the batch size and the number of visual tokens yields minimal improvement. In contrast, increasing the number of resampling layers significantly reduces the loss without substantially increasing the training budget.

\noindent \textbf{Effects of Scaling LLM Parameter Size (70B \textit{vs.} 13B)}:
The 70B model demonstrates a reduced training loss overall when compared with the 13B versions.
This loss performance also aligns with the downstream task results in Tables \ref{tab:exp:image_captioning} and \ref{tab:exp:vqa}.
We attribute this result to the inherent reasoning abilities and the knowledge assimilated within the 70B models, which expedites the visual concept acquisition and alignment process.
Overall, the comparison demonstrates the importance of scaling LLM parameters in vision-language pre-training as well, which is an aspect that has seldom been addressed in existing literature.

\begin{table*}[t]
\centering
\scalebox{0.95}{
    \begin{tabular}{p{3.5cm}p{4cm}p{6cm}}
    \toprule
    \textbf{IMU Signals Input} & \textbf{Reference Video} & \textbf{\mn Output} \\
    \toprule
    \multirow{4}{*}{        
        \begin{minipage}{3.5cm}
            \includegraphics[width=3.4cm]{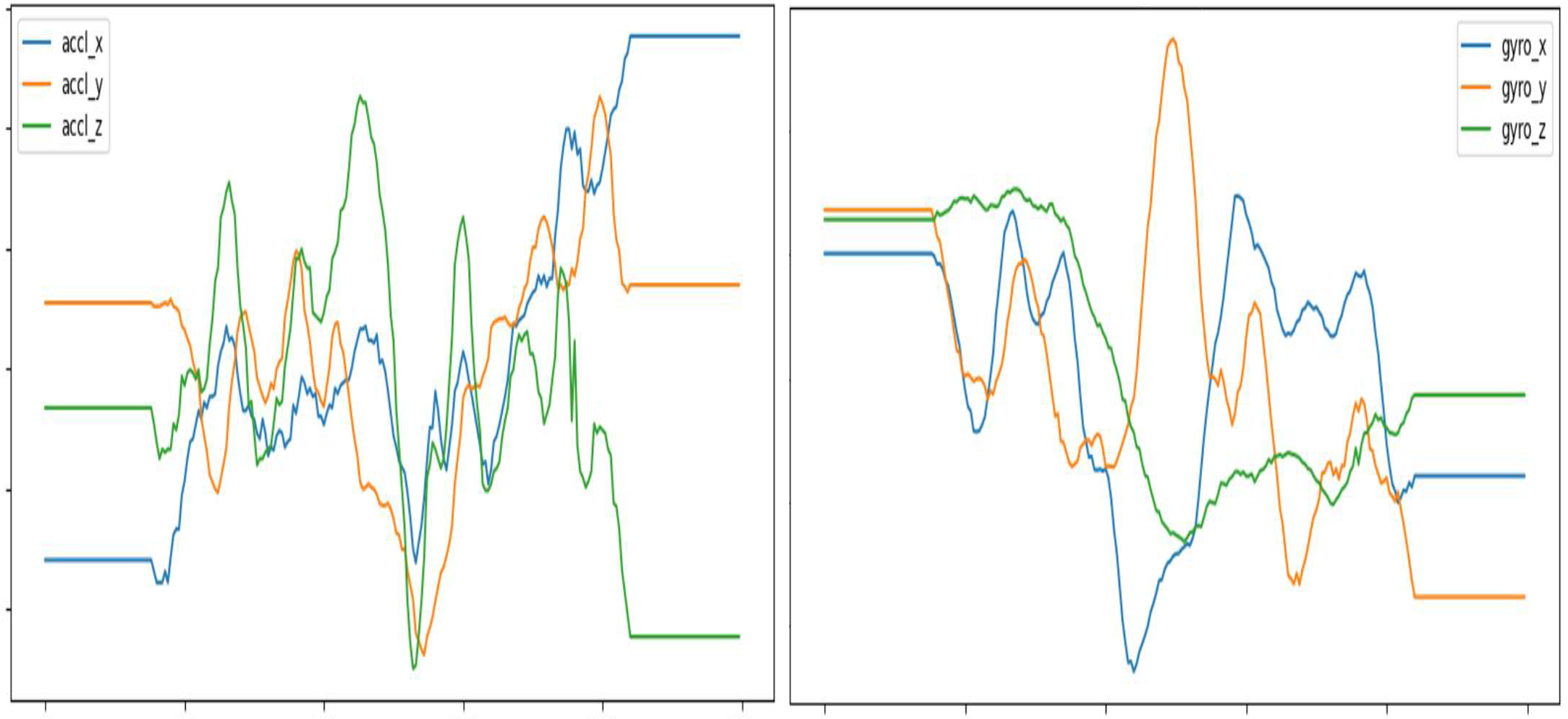}
        \end{minipage}
    }
    &
    \multirow{4}{*}{        
        \begin{minipage}{4.0cm}
            \includegraphics[width=3.9cm]{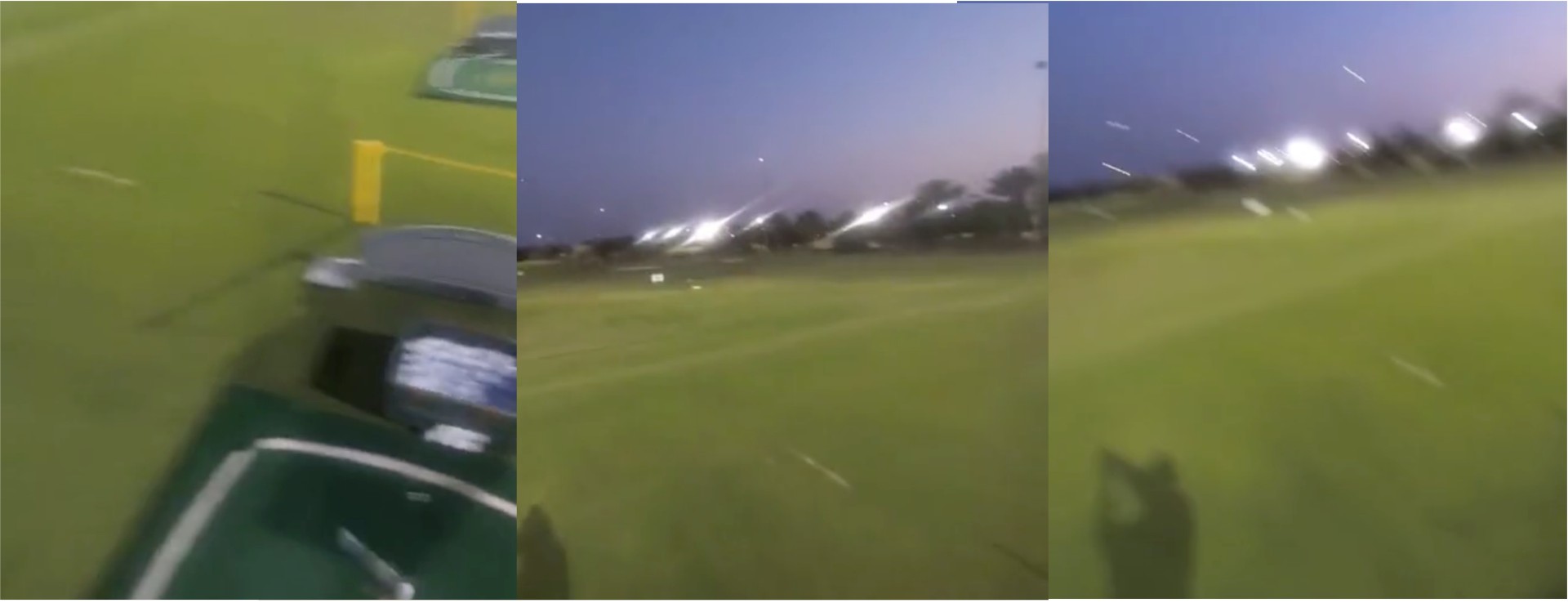}
        \end{minipage}
    }    
    & \textbf{Prompt}: Describe the motion. \\
    & & \textbf{Response}: [Camera wearer] looks around. \\
    & & \\
    & & \\

    \midrule
    \multirow{2}{*}{        
        \begin{minipage}{3.5cm}
            \includegraphics[width=3.4cm]{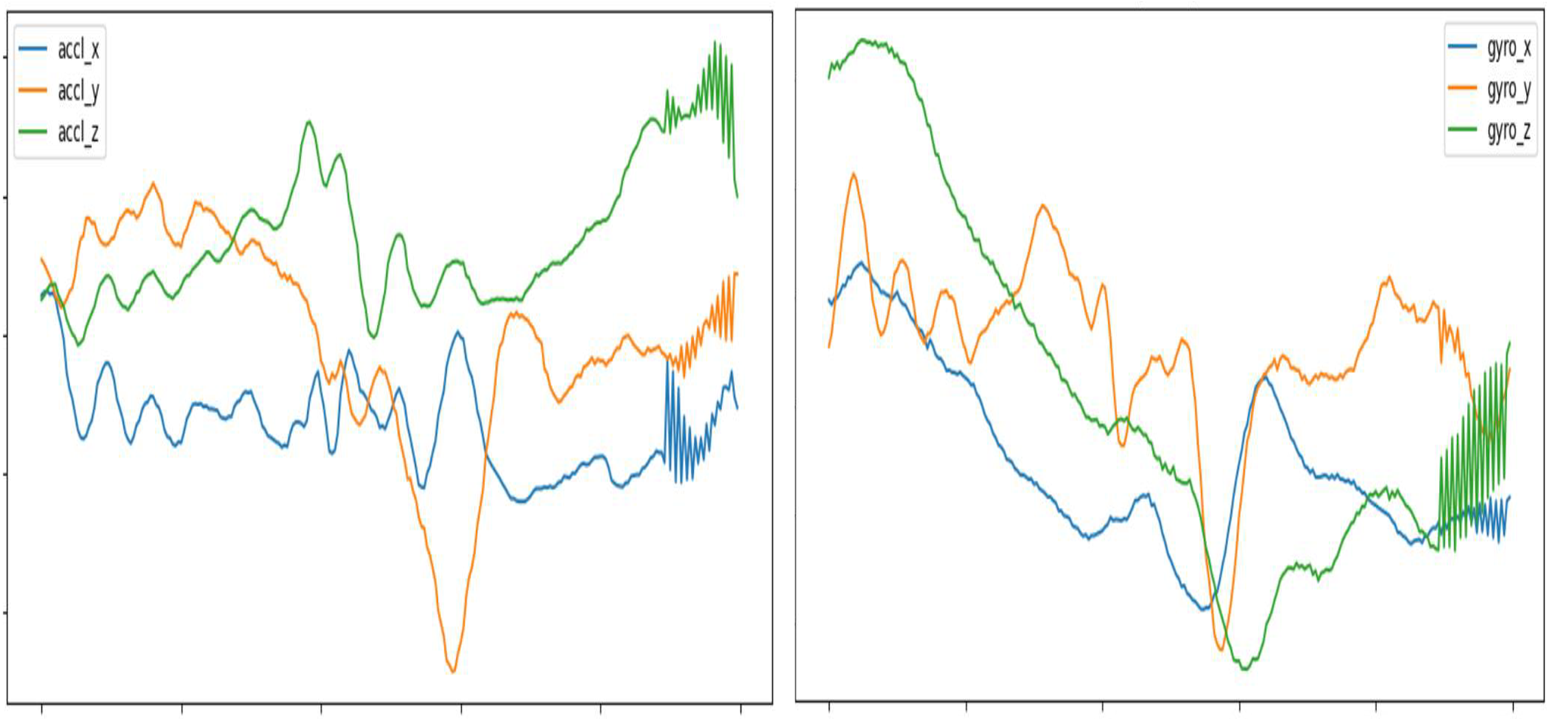}
        \end{minipage}
    }
    &
    \multirow{2}{*}{        
        \begin{minipage}{4.0cm}
            \includegraphics[width=3.9cm]{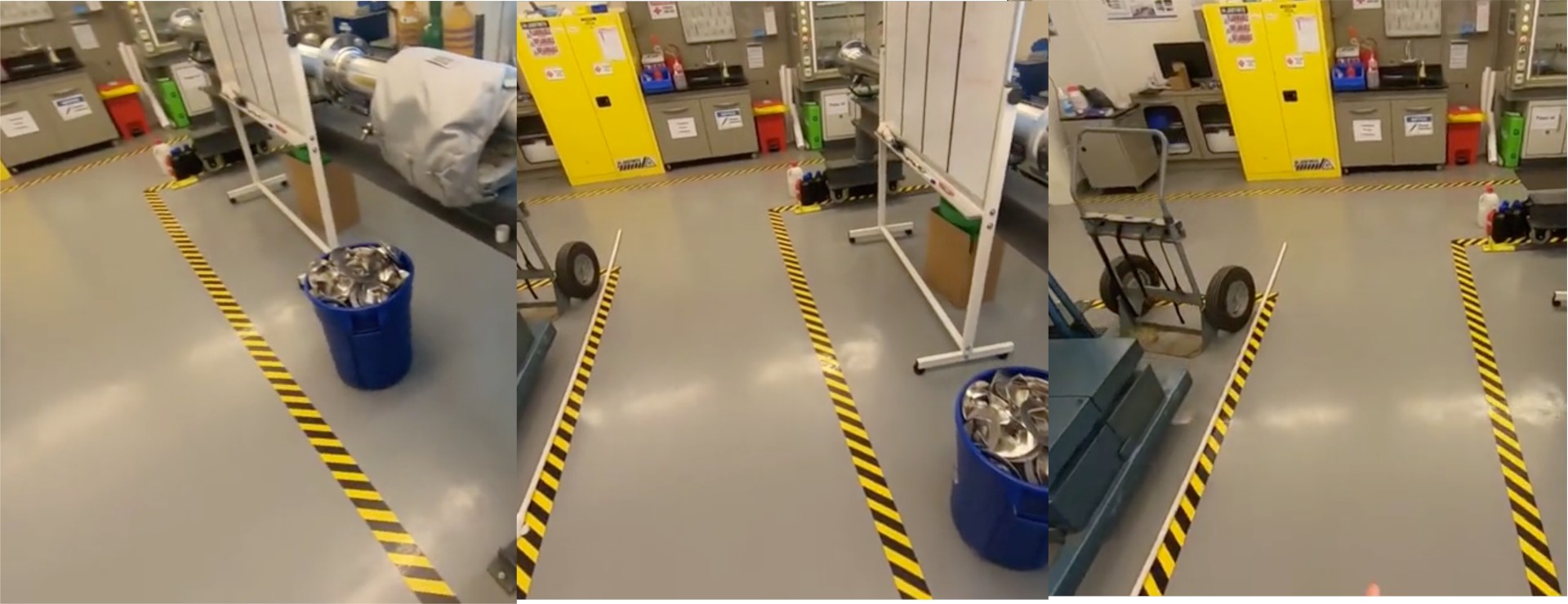}
        \end{minipage}
    }    
    & \textbf{Prompt}: Describe the motion. \\
    & & \textbf{Response}: [Camera wearer] walks in the house. \\
    & & \\

    \midrule
    \multirow{2}{*}{        
        \begin{minipage}{3.5cm}
            \includegraphics[width=3.4cm]{figures/examples/imu_signal_1.png}
        \end{minipage}
    }
    &
    \multirow{2}{*}{        
        \begin{minipage}{4.0cm}
            \includegraphics[width=3.9cm]{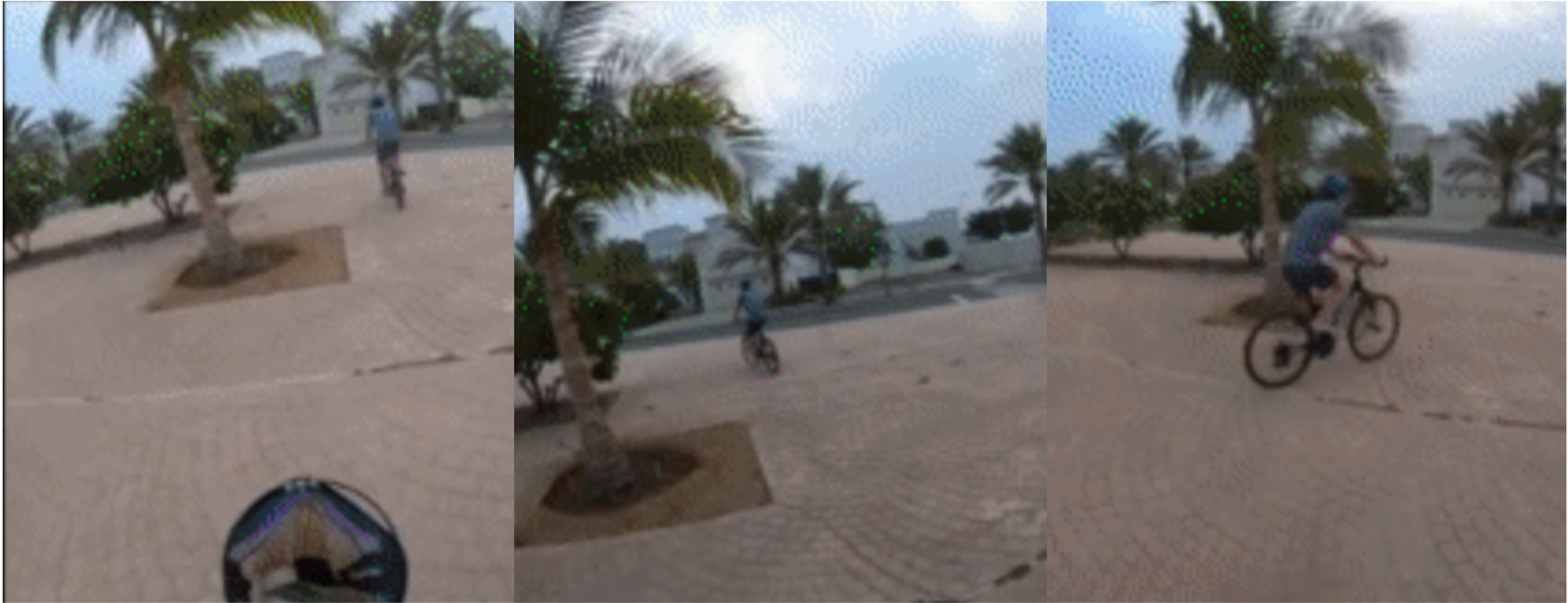}
        \end{minipage}
    }    
    & \textbf{Prompt}: What's the safest way to stop? \\
    & & \textbf{Response}: To stop safely on a bike, use the brakes gradually and look where you want to go. Lean the bike slightly to one side and keep your body upright. \\
    & & \\

    \bottomrule
    \end{tabular}
}
\caption{
Example outputs from \mn, on \underline{IMU motion sensor} samples from Ego4D. The text prompt and the IMU signals are used as input. We display a few frames from the synchronized video as a reference for readers. The reference video is not used as input to the model.
}
\label{tab:imu_examples}
\end{table*}

\noindent \textbf{Audio Caption Generation}: 
Table \ref{tab:aac} shows the audio captioning results on the AudioCaps \cite{kim2019audiocaps} benchmark dataset. 
\mn significantly outperforms other state-of-the-art audio captioning models in the literature (\eg +10.9pp in CIDEr, +5.8pp in SPICE), showing the versatility of the proposed approach on various modalities beyond just vision.
We note that our 70B model displays notably strong performance compared to the 7B and the 13B variants -- showing the importance of the reasoning module for the task.


\noindent \textbf{IMU Motion Description Generation}: 
We use the Ego4D \cite{ego4d} dataset to train an IMU-aligned \mn-7B model, leveraging the synchronized IMU sensor data and textual narrations provided in the dataset.
Given that the task of generating textual descriptions from motion signals has not been previously achievable or reported, we solely present the performance achieved by our own model.

\begin{table*}[!ht]
\centering
\scalebox{0.97}{
    \begin{tabular}{p{2.2cm}p{11.5cm}}
    \toprule
    \multirow{2}{*}{\textbf{Example 1. Creative Writing}} \\ \\
    \midrule
    \midrule
    \textbf{Input Image} &
        \begin{minipage}{.6\textwidth}
            \includegraphics[height=40mm]{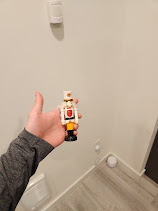}
            \vspace{2pt}
        \end{minipage}
    \\
    \textbf{Prompt} & \textbf{\textit{Write a joke about it.}} \\
    \midrule    
    BLIP-2 & a person holding a lego toy in their hand \\
    \cmidrule(r){2-2}
    InstructBLIP & a hand holding a small toy in a room \\
    \cmidrule(r){2-2}    
    Mini-GPT4 & Why did the lego man break up with his lego wife? Because he found a better lego man. \\
    \cmidrule(r){2-2}
    OpenFlamingo & What do you get when you cross a snowman with a robot? Answer: A snowbot. \\
    \cmidrule(r){2-2}
    LLaVA & A British solider mannequin has lost its way in a New York City apartment. When the surprised tenant, who doesn't seem to know much about British (...) \\
    \midrule
    \mn 70B & Here's a joke for you: Why did the nutcracker doll go to the doctor? Because he was feeling a little cracked!  \\

    \toprule
    \multirow{2}{*}{\textbf{Example 2. How-To}} \\ \\
    \midrule
    \midrule
    \textbf{Input Image} &
        \begin{minipage}{.6\textwidth}
            \includegraphics[height=40mm]{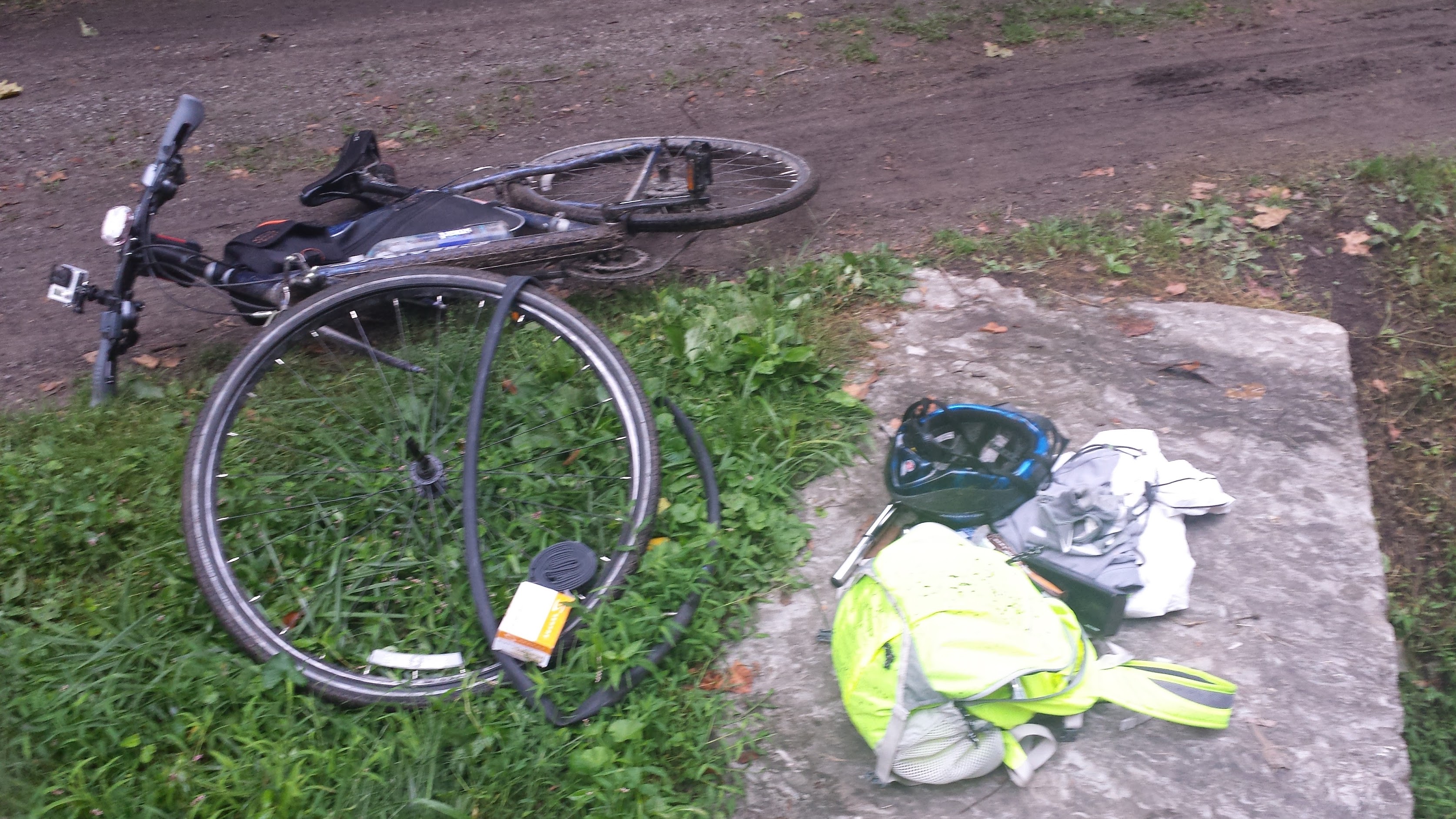}
            \vspace{2pt}
        \end{minipage}
    \\
    \textbf{Prompt} & \textbf{\textit{How do I fix it?}} \\
    \midrule    
    BLIP-2 & a bike that has been thrown off a cliff and landed on a rock in the middle of a field  \\
    \cmidrule(r){2-2}
    InstructBLIP & tire \\
    \cmidrule(r){2-2}    
    Mini-GPT4 & The bike has a flat tire. To fix it, you will need to pump up the tire with air. First, remove the wheel from the bike and place it on the ground. Then,  (...)  \\  
    \cmidrule(r){2-2}
    OpenFlamingo & This is a flat tire. \\
    \cmidrule(r){2-2}
    LLaVA & It appears that the person biking along the way has experienced a flat tire because their bike is leaning against a wall, and there is a bike tube lying on the ground. To fix the issue, the person would need to first examine the tire to determine the location of the flat. They should then remove the damaged tube from the tire. (...) \\
    \midrule
    \mn 70B &  It appears that the front tire of one bicycle is flat. You might want to try changing the inner tube or patching any holes in the existing one before re-inflating the tire. \\
    \bottomrule
    \end{tabular}
}
\caption{
    Examples comparing \mn with various baselines: BLIP-2 (FlanT5$\_$XXL) \cite{blip2}, InstructBLIP (Vicuna-13B) \cite{instructblip}, OpenFlamingo-v2-9B \cite{openflamingo}, Mini-GPT4 \cite{minigpt4}, and LLaVA \cite{llava}.
    \mn demonstrates competitive performance in object recognition (\eg ``nutcracker''), response relevancy (\eg writing a joke, explaining how-to), and secondary reasoning. 
}
\label{tab:qual_examples_1}
\end{table*}

\begin{table*}[!ht]
\centering
\scalebox{0.97}{
    \begin{tabular}{p{2.2cm}p{11.5cm}}
    \toprule
    \multirow{2}{*}{\textbf{Example 3. Recommendation}} \\ \\
    \midrule
    \midrule
    \textbf{Input Image} &
        \begin{minipage}{.6\textwidth}
            \includegraphics[height=50mm]{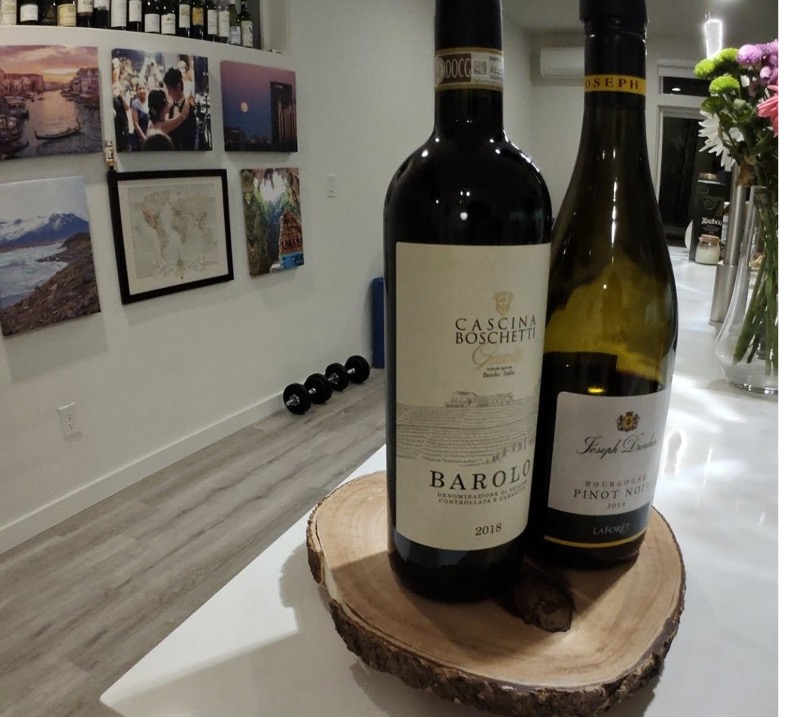}
            \vspace{2pt}
        \end{minipage}
    \\
    \textbf{Prompt} & \textbf{\textit{Which one would pair better with steak?}} \\
    \midrule    
    BLIP-2 & red wine or white wine? \\
    \cmidrule(r){2-2}    
    InstructBLIP & red 1 \\
    \cmidrule(r){2-2}
    Mini-GPT4 & The wine bottles in the image are two different types of wine, one is a red wine and the other is a white wine. Both of these wines would pair well with steak. The red wine would be a good choice for a rich and flavorful steak, while the white wine would be a good choice for a lighter steak or seafood dish. \\
    \cmidrule(r){2-2}
    \cmidrule(r){2-2}
    LLaVA & For a steak, the red wine is likely a more appropriate pairing compared to the white wine. Whether it is Chianti or Chianti Classico, both options are Italian red wines made from Sangiovese grapes, and their full-bodied nature provides a rich flavor profile, making them good companions to steak dishes. The white wine, Soave Classico DOC, is a white grape varietal, which is less traditional for pairing (...) \\
    \midrule
    \mn 70B & The Barolo on the left would pair better with steak. It's a full-bodied, high tannin wine that can stand up to the richness of red meat.  \\

    \toprule
    \multirow{2}{*}{\textbf{Example 4. Question and Answering}} \\ \\
    \midrule
    \midrule
    \textbf{Input Image} &
        \begin{minipage}{.6\textwidth}
            \includegraphics[height=37mm]{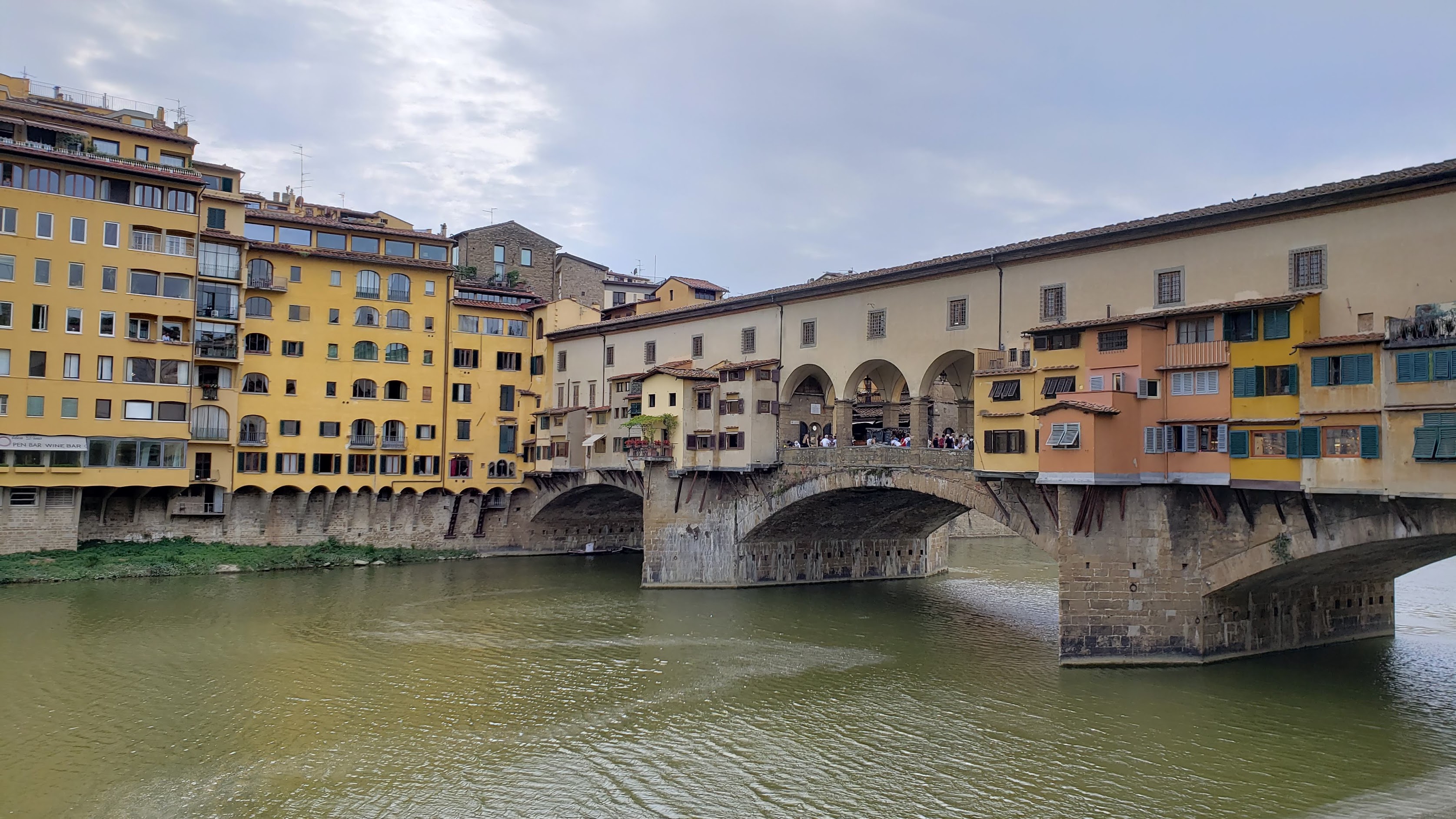}
            \vspace{2pt}
        \end{minipage}
    \\
    \textbf{Prompt} & \textbf{\textit{How long does this river go?}} \\
    \midrule    
    BLIP-2 & it's a river that runs through the city of florence, tuscany, italy  \\
    \cmidrule(r){2-2}
    InstructBLIP & 10 miles  \\
    \cmidrule(r){2-2}
    Mini-GPT4 & The river goes over the bridge in the image. \\
    \cmidrule(r){2-2}
    \cmidrule(r){2-2}
    LLaVA & The river in the image is quite long, stretching across the city and passing under several buildings, including a bridge. \\
    \midrule
    \mn 70B &  The river in the image is the Arno River, which flows through Florence, Italy. It stretches for approximately 241 kilometers (150 miles) and empties into the Ligurian Sea.\\
    \bottomrule
    \end{tabular}
}
\caption{
    Examples comparing \mn with various baselines.
    \mn demonstrates competitive performance in both object recognition (\eg ``Barolo'', ``Arno River'') and secondary reasoning. 
}
\label{tab:qual_examples_2}
\end{table*}

On the held-out test set, we achieve 52.5 CIDEr and 23.2 ROUGE-L against the ground-truth captions, showing the feasibility of the newly proposed task.

Combining this captioning ability with the reasoning capabilities of LLMs, in Table \ref{tab:imu_examples} we show examples of \textit{novel} applications that \mn might allow, \eg inferring user motion states and incorporating these as part of its response (\eg ``\textit{What's the safest way to stop?}''$\rightarrow$``\textit{To stop safely on a bike, ...}'' without any textual or visual cues that the user is biking).

\begin{table*}[t]
\centering
\scalebox{0.95}{
    \begin{tabular}{p{3.5cm}p{4cm}p{6cm}}
    \toprule
    \textbf{Input Modality 1} & \textbf{Input Modality 2} & \textbf{\mn Output} \\
    \toprule
    \multirow{4}{*}{        
        \begin{minipage}{3.5cm}
            \includegraphics[width=3.4cm]{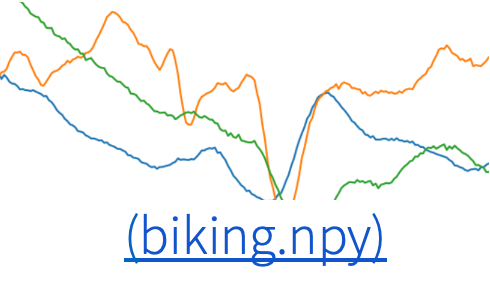}
        \end{minipage}
    }
    &
    \multirow{4}{*}{        
        \begin{minipage}{4.0cm}
            \includegraphics[width=3.9cm]{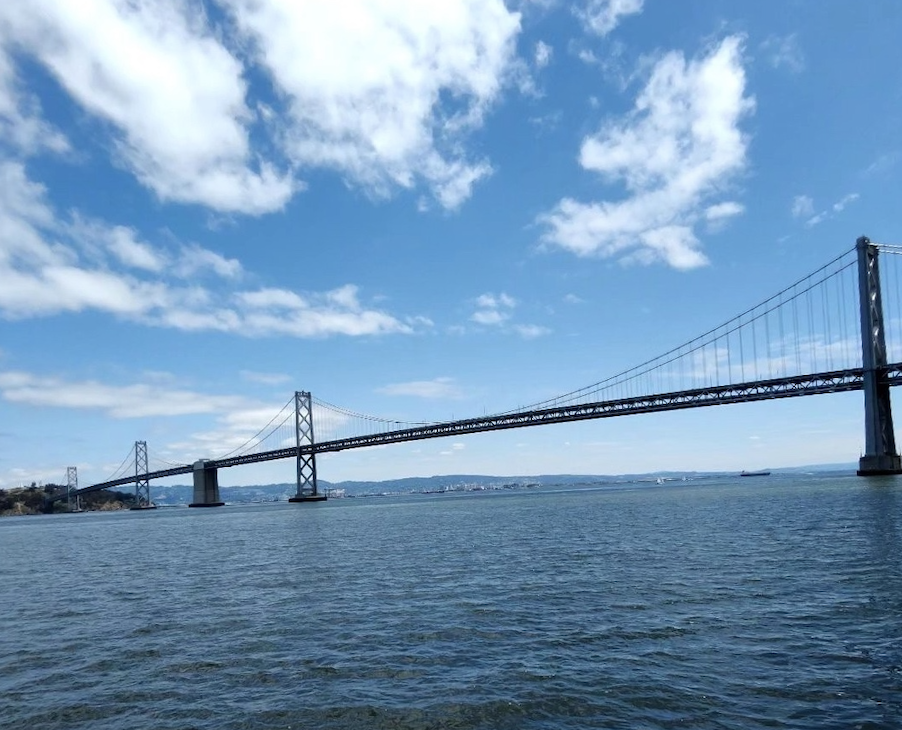}
        \end{minipage}
    }    
    & \textbf{Prompt}: (Given the motion signals)  Write a social media caption for this view. \\
    & & \\    
    & & \textbf{Response}: \underline{Pedaling} along the San Francisco Bay, taking in breathtaking views of the \underline{Golden Gate Bridge}! The salty sea breeze invigorates me as I make my way to my next destination. \\

    & & \\

    \bottomrule
    \end{tabular}
}
\caption{
Example outputs from \mn, with multiple \textbf{interleaved modalities} as input. The text prompt and two other modalities (\eg image \& IMU motion sensor signals) are used as input.
The \underline{underlined} text in the response demonstrates the evidence that the output is grounded on multiple modalities.
}
\label{tab:interleaved_examples}
\end{table*}

\subsection{Qualitative Analysis}
\label{subsec:experiments:qualitative}

\noindent \textbf{Comparison with other Vision-Language Models}:
Table \ref{tab:qual_examples_1} and \ref{tab:qual_examples_2} show outputs from various vision-language models \cite{instructblip, minigpt4, openflamingo, llava} on diverse example image and prompt pairs, compared with \mn (LLaVA-70B).
It can be seen that \mn exhibits strong visual understanding capabilities (such as identification of objects and their states), as well as language generation capabilities.
While MiniGPT4 \cite{minigpt4} and LLaVA \cite{llava} do present reasonable and fluent responses, their accuracy is not consistently ensured.
These examples effectively highlight the benefits of the proposed approach which allows for large-scale pre-training covering diverse visual concepts, while inheriting strong reasoning capabilities derived from instruction-tuned LLMs.

We note that we use the latest checkpoints made available for each baseline to generate responses.

\noindent \textbf{Interleaved Modalities}: 
The flexible model architecture of \mn allows for combinatory modalities as conditioning context (\eg image + IMU motion sensor signals), which allows for more comprehensive multimodal reasoning.
We demonstrate the model's zeroshot capabilities of handling such interleaved modalities in Table \ref{tab:interleaved_examples} (\eg composing a message with a given image (Golden Gate Bridge), with the user's prevalent motion (biking) as part of the context).

This result illustrates the new and natural way of interaction with an AI model made possible by \mn, wherein a user can presume a shared understanding of combined sensory perceptions (\eg visual, auditory, and motion cues) when composing queries -- avoiding the need to specify multimodal contexts.

\section{Safety}
\label{sec:safety}


\noindent \textbf{Inference Time Integrity}.
To ensure the safety and integrity of the \mn model, several measures are made on the following categories of potential integrity violations: (1) input images, (2) input text prompts, (3) text outputs, and (4) multimodal combination of input images and text outputs.

\begin{enumerate}[label=(\arabic*)]
    \item Input image: we use a pre-trained image classifier based on RegNetY \cite{radosavovic2020designing} to detect any content that violates integrity standards.  This detection encompasses graphic material, violent imagery, hate symbols, instances of bullying, harassment, \etc. If such a violation is identified within the image, we proceed to reject the entire query.
    \item Input text prompt: we use a RoBERTa-based text classifier \cite{roberta} trained to detect integrity-violating utterances such as violence, harassment, hate speech, \etc. When a violation is detected in user prompt, we proceed to reject the entire query.
    \item Output text: we employ the same text classifier in (b) to detect any problems within the generated output. For streaming use cases, we run the classifier for each sentence to promptly identify any violations.
    \item Multimodal association in input image \& output text: in the uncommon scenario where harmless text and a harmless image (which individually appear innocuous) can result in a problem when they are associated, we use a multimodal classifier to detect such instances.
\end{enumerate}

\noindent \textbf{Training Time Safety}.
The datasets used for pre-training (e.g. \cite{laionfiltering,singer2022make}) have gone through a filtration process to remove harmful language or images that compromise integrity, thereby reducing the potential for the model to generate content that violates integrity standards.

\noindent \textbf{LLM Safety}.
Since our \mn pre-training does not alter the parameters of the base LLM, we carry over the same safety precautions implemented for its language generation.
For instance, LLaMA-2 (the version we report most of our results on) places safeguards such as negative example fine-tuning, reinforcement learning with human feedback (RLHF) \cite{rlhf, bai2022constitutional, dpo}.

\section{Conclusions}
\label{sec:conclusions}

Our proposed \mn showcases a novel and natural way of interacting with an AI model, \eg asking questions that presume a shared understanding of the world between the user and the agent, through the same lens and combinatory perceptions (\eg visual, auditory, and motion cues).
The proposed scalable way of training \mn makes it possible to leverage the powerful reasoning capabilities of the LLaMA-2 language model within the multimodal settings.

Our contributions are as follows:
(1) We present a large-scale Multimodal LLM (\mn), trained using open-sourced resources and scalable solutions for multiple modalities.
(2) We introduce the \dnf{} dataset (\dn), a first-of-its-kind collection of high-quality manual annotations of multimodal instruction data.
(3) Our comprehensive empirical analysis shows insights to the efficient and scalable \textit{recipe} for building a multimodal reasoning model, given various LLMs and modeling choices.

\section{Limitations}
We discuss the current limitations of our work as follows.

First, the proposed causal multimodal language modeling approach still encounters challenges in establishing a robust grounding with the input modality.
Specifically, we observe that during the generation, the model occasionally prioritizes focusing more on the generated text rather than the input image. This leads to the generation of output that incorporates biases acquired from the underlying language model (LLM), which can incur inaccuracies when compared against the image context.
We expect that additional architectural adjustments or unfreezing LLM parameters are necessary to address this limitation effectively (albeit the much higher computational costs it might entail).

Second, while we greatly increase the size of the pretraining dataset, the understanding of visual concepts and entities remains constrained by the quantity of paired image-text data included in the training process.
In the domain of text-only language models, it is commonly observed that approaches incorporating external knowledge retrieval significantly enhance the model's ability to overcome its knowledge limitations. These approaches offer a potential means to alleviate the limitations mentioned earlier.

Lastly, in the scope of our work, the multimodal adaptation of an LLM is bounded by four modalities: image, video, audio, and IMU signals. 
While we believe that the proposed approach has the potential to encompass any other modality, provided there exists a paired dataset, its effectiveness for such modalities still needs to be substantiated.


\bibliographystyle{ieeetr}

\clearpage

\appendix

\section{Human Evaluation}
\label{a-sec:human_eval}

To measure if the \mn response is relevant and correct with respect to the query and the image, we use the following rubrics to collect human feedback data.

\noindent \textbf{Recognition Accuracy}: Did the model correctly understand and identify the objects or scenarios in the image relevant to the query?

\begin{itemize}
    \item 0: The model completely misinterprets the image or fails to recognize key objects in the image relevant to the query.
    \item 1: The model partially misinterprets the image or recognizes some but not all key elements.
    \item 2: The model correctly interprets the image and identifies key elements.
\end{itemize}

\noindent \textbf{Response Accuracy}: Does the response provide factually correct information? Does it Hallucinate?

\begin{itemize}
    \item 0: The response contains factually incorrect information or even partially incorrect information.
    \item 2: The response provides factually accurate information.
\end{itemize}

\noindent \textbf{Integrity}: Does the response show any harmful or offensive language or signs of unjust or prejudiced treatment related to characteristics such as race, age, gender, nationality, etc.?

\begin{itemize}
    \item 0: The response shows some sign of bias or harmful/offensive language.
    \item 2: The response shows no signs of bias or harmful/offensive language.
\end{itemize}

\noindent \textbf{Pairwise Comparison}: choose the response that better answers the given question, and that is more contextually appropriate and accurate (win, tie, or lose).

We plan to explore further training the model with the human feedback data (\eg RLHF). 

\section{Additional Notes on Experiments}


Different prompts were used to get the model output in the desired format for each task (\eg multiple choice questions, yes/no questions).
Below is the full list of prompts used for each task.

\subsection{Multimodal Prompts}
\label{a-sec:prompts}
\paragraph{MM-IT} System message: ``\textit{You are a multimodal assistant, designed to provide helpful answers to users' image-related questions. $\backslash$n$\backslash$n Here is the image: <img>}". User message: ``\textit{\{question\}"}

\paragraph{VQA, TextVQA, OKVQA} System message: ``\textit{You are a multimodal assistant, designed to provide direct answers to users' image-related questions. Reply directly with only one phrase. *Do not* start your answer with `Sure ...'. $\backslash$n$\backslash$n Here is the image: <img>}". User message: ``\textit{In the image, \{question\} Reply in one word.}

\paragraph{VizWiz} System message: ``\textit{Answer the questions based on the image when possible, otherwise say `unanswerable`. $\backslash$n$\backslash$n Here is the image: <img>}". User message: ``\textit{In the image, \{question\} Reply in one prahse/word or say `unanswerable`}

\paragraph{Hateful Meme} System message: ``\textit{You are a social media content moderator, designed to detect hateful memes. $\backslash$n$\backslash$n Here is the meme: <img>$\backslash$n This meme contains text: `\{ocr\}'}". User message: ``\textit{Is this a hateful meme? Answer yes or no.}

\paragraph{Coco Caption} System message: ``\textit{You are a multimodal assistant, designed to provide direct and concise answers to users' image-related requests. $\backslash$n$\backslash$n Here is the image: <img>}". User message: ``\textit{Describe the image with one *generic* sentence using json format. Here are two examples:$\backslash$n Specific: \{"caption": "Body-Solid (Best Fitness) Inversion Table-2"\} $\backslash$n Generic: \{"caption": "A man laying on top of an exercise table.\}.}

\paragraph{ScienceQA} System message: ``\textit{Given the image, choose the correct option for the following question. Your response must be just a single letter that corresponds to the correct option (e.g. A, B) $\backslash$n$\backslash$n Here is the image: <img>." User message: ``\{context\} Question: \{question\} $\backslash$n$\backslash$n Options: \{choices\} $\backslash$n$\backslash$n Reply in a single letter.}"

\paragraph{AudioCap} System message: ``\textit{You are a multimodal assistant. Designed to provide direct answers to users' audio-related questions. Here is the audio: <audio>}" User message: ``\textit{Describe the sound.}"

\paragraph{STAR, How2QA, NextQA} System message: ``\textit{You are a multimodal assistant. Designed to provide direct answers to users' video-related questions. $\backslash$n$\backslash$n Here is the video: <video>.}" User message: \{question\} Select exactly one option from the following: [{options}].

\paragraph{IMU-Ego4d} System message: ``\textit{"You are a multimodal assistant, designed to provide helpful, concise and direct answers to users' questions, based on the user's motion sensor signals reading from a head-mounted IMU device. The signals may indicate that a user may be running, walking, biking, driving, looking around, etc. Always answer under 30 words. $\backslash$n$\backslash$n Here are the user's predicted motions: <IMU>"} User message: ``\textit{Describe this motion.}"

\subsection{Multimodal Inputs}
\begin{figure}[t]
    \centering
    \includegraphics[width=0.8\columnwidth]{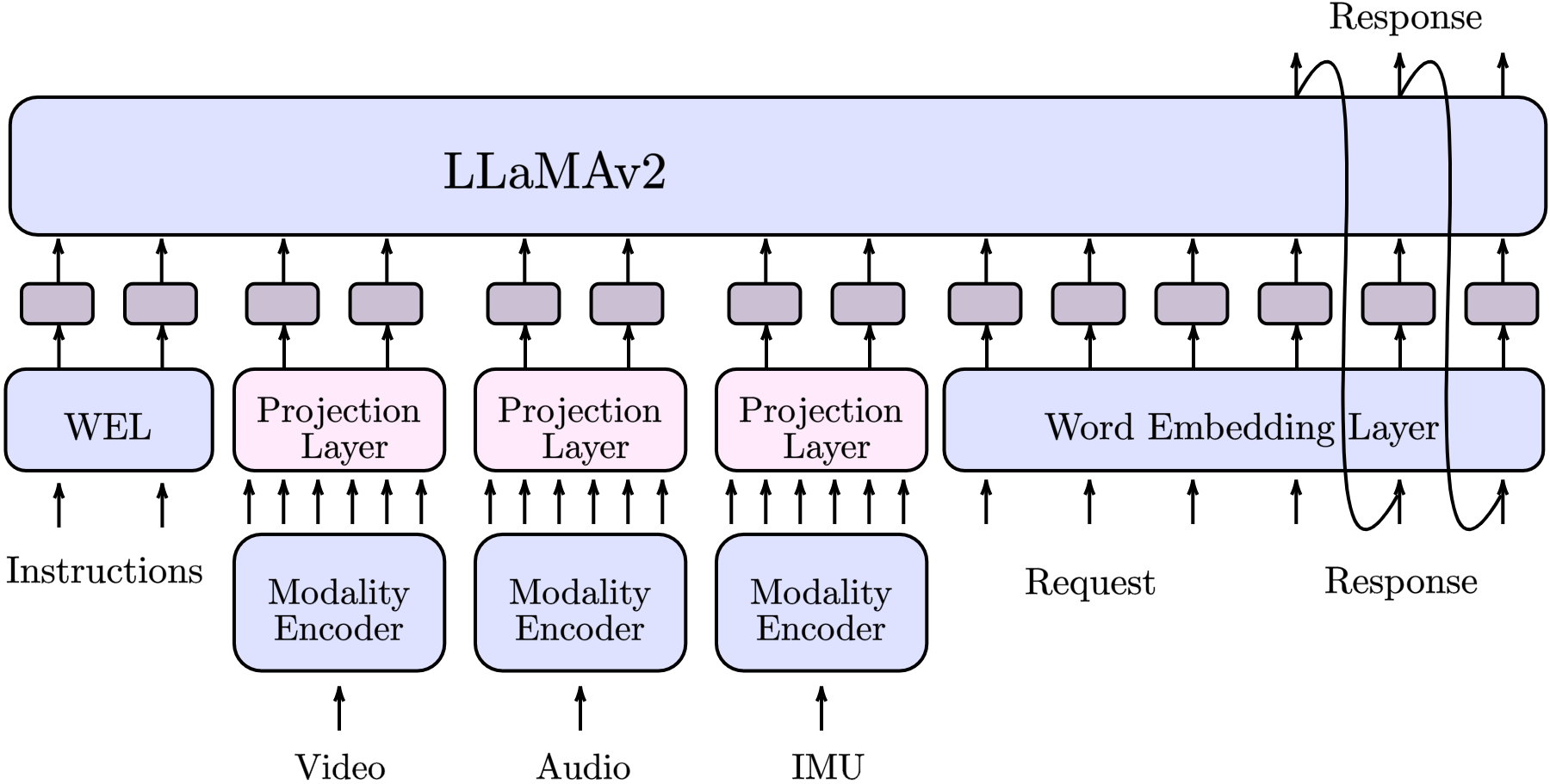}
    \caption{
    \mn Inference example with multiple modality as input.
    }
    \vspace{-15pt}
    \label{fig:diagram_MM}
\end{figure}

Figure \ref{fig:diagram_MM} shows the diagram for performing modality-interleaved inference (for examples shown in Table \ref{tab:interleaved_examples}).

\subsection{Hyperparameters}
\label{a-sec:hparams}

\paragraph{Pre-training:} Table \ref{tab:hyparams} report the hyper-parameters used in this work for model pre-training.

\begin{table*}[th!]
\centering
\footnotesize
\centering
\begin{tabular}{cccccccc}
    \toprule
    \multicolumn{1}{c}{\multirow{2}{*}{\textbf{Models}}} & \multirow{2}{*}{\textbf{Batch Size}} & \multirow{2}{*}{\textbf{Initial LR}} & \multirow{2}{*}{\textbf{\# Steps}} & \textbf{\# Modality} & \multirow{2}{*}{\textbf{Projection Module (\#Layers)}}  \\
    & & & & \textbf{Embeddings} & \\
    \midrule
    \multicolumn{1}{l}{\mn (13B, Image)}  & 2048 & $2 \times 10^{-4}$ & 100k  & 64 & Resampler (6) \\
    \multicolumn{1}{l}{\mn (70B, Image)}  & 2048 & $2 \times 10^{-4}$ & 100k  & 64 & Resampler (6) \\
    \multicolumn{1}{l}{\mn (13B, Audio)}  & 128 & $1 \times 10^{-4}$ & 1k  & 32 & Linear (1) \\
    \multicolumn{1}{l}{\mn (70B, Audio)}  & 128 & $1 \times 10^{-4}$ & 1k  & 32 & Linear (1) \\
    \multicolumn{1}{l}{\mn (13B, Video)}  & 1024 & $1 \times 10^{-4}$ & 20k  & 32 & Resampler (4) \\
    \multicolumn{1}{l}{\mn (70B, Video)}  & 1024 & $1 \times 10^{-4}$ & 20k  & 32 & Resampler (4) \\
    \multicolumn{1}{l}{\mn (7B, IMU)}  & 256 & $1 \times 10^{-4}$ & 2k  & 32 & Linear (1) \\
    \bottomrule
\end{tabular}
\caption{
    \footnotesize Hyperparameters for AnyMAL Pre-training
}
\label{tab:hyparams}
\end{table*}

\paragraph{Finetuning:} We finetune both resampler and LoRA adapters on MM-IT training set with the prompt described in \ref{a-sec:prompts}. Following the QLoRA \cite{qlora} work, we set LoRA $r = 64$, $\alpha = 16$, and add LoRA modules on all linear layers. We finetune the model for 3k steps with batzh size 128, initial LR $1 \times 10^{-5}$.

\subsection{Code Base \& Hardware}
The implementations of the transformer-based models are extended from the HuggingFace\footnote{https://github.com/huggingface/transformers}~code base~\cite{transformers} and other cited authors' released code-bases.
Our entire code-base is implemented in PyTorch \cite{pytorch}. 
All models in this work are trained on a varying number of Nvidia A100 \footnote{https://www.nvidia.com/en-us/data-center/a100/} GPUs.

\subsection{More Qualitative Examples}

We provide more qualitative examples in Table \ref{tab:qual_examples_3} and \ref{tab:qual_examples_4}, comparing outputs from various baselines.
While some of the baselines do present reasonable and fluent responses, their accuracy is not consistently ensured, either in their visual understanding (\eg what objects are present in an image) or secondary reasoning (\eg analyzing whether there is any element in an image that violates integrity policy).

\begin{table*}[t]
\centering
\scalebox{0.95}{
    \begin{tabular}{p{2.7cm}p{11.0cm}}
    \toprule
    \multirow{2}{*}{\textbf{Example 5. Recommendation - Fashion}} \\ \\
    \midrule
    \midrule
    \textbf{Input Image} &
        \begin{minipage}{.6\textwidth}
            \includegraphics[height=40mm]{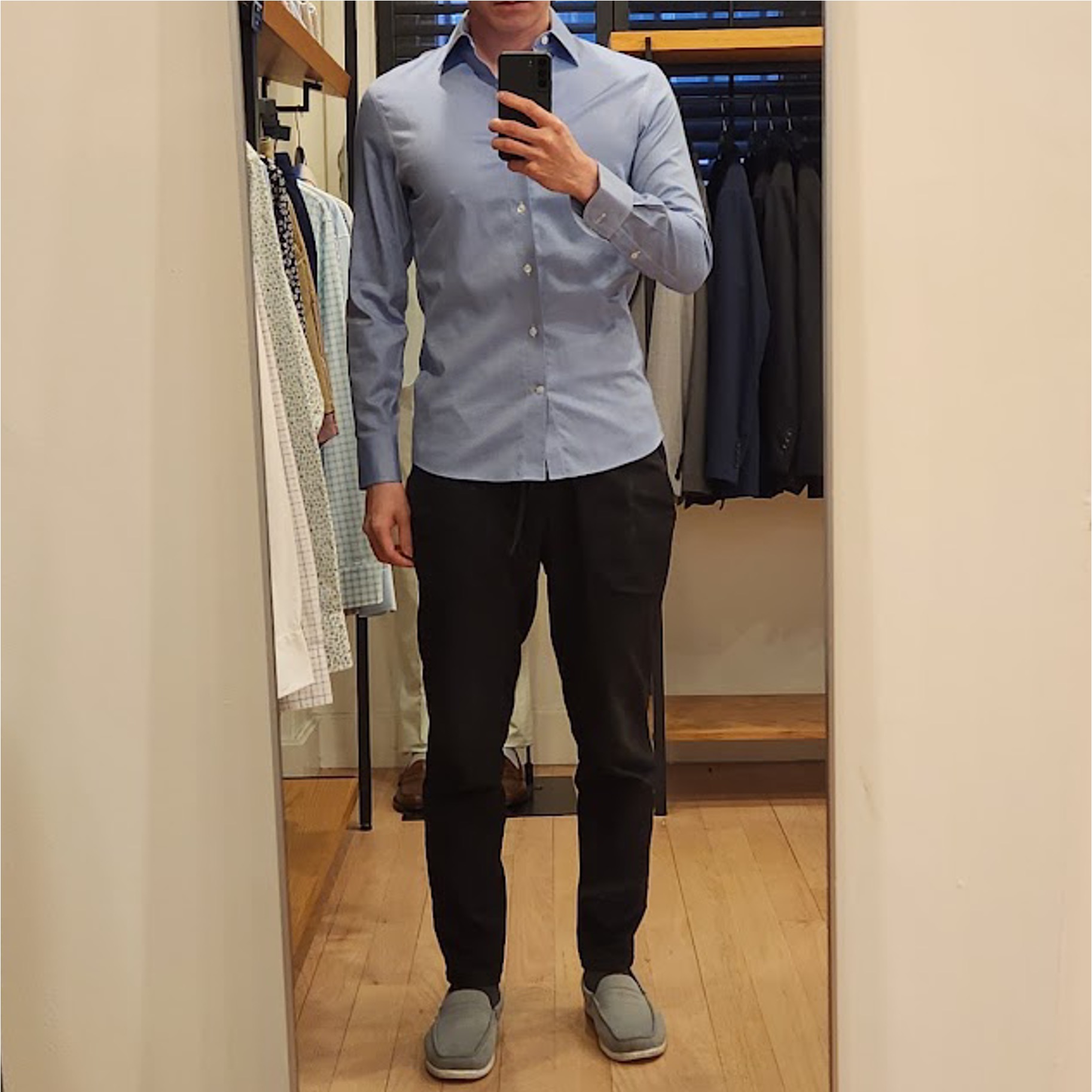}
            \vspace{2pt}
        \end{minipage}
    \\
    \textbf{Prompt} & \textbf{\textit{What jacket would go well with this?}} \\
    \midrule    
    BLIP-2 & blue shirt, black pants, grey shoes \\
    \cmidrule(r){2-2}    
    InstructBLIP & black \\
    \cmidrule(r){2-2}
    Mini-GPT4 & A light blue shirt would go well with this outfit. \\
    \cmidrule(r){2-2}
    OpenFlamingo & This jacket would go well with this outfit. \\
    \cmidrule(r){2-2}
    LLaVA & With the man wearing a blue shirt and a pair of black pants, a solid-colored blazer, such as a navy or black one, would go well with the outfit. \\
    \midrule
    \mn 70B & A navy blazer would look great with this shirt. The navy color will complement the shirt's blue hue, and the blazer will add a more formal touch to the outfit.  \\

    \toprule
    \multirow{2}{*}{\textbf{Example 6. Multimodal Integrity Violation Detection}} \\ \\
    \midrule
    \midrule
    \textbf{Input Image} &
        \begin{minipage}{.6\textwidth}
            \includegraphics[height=40mm]{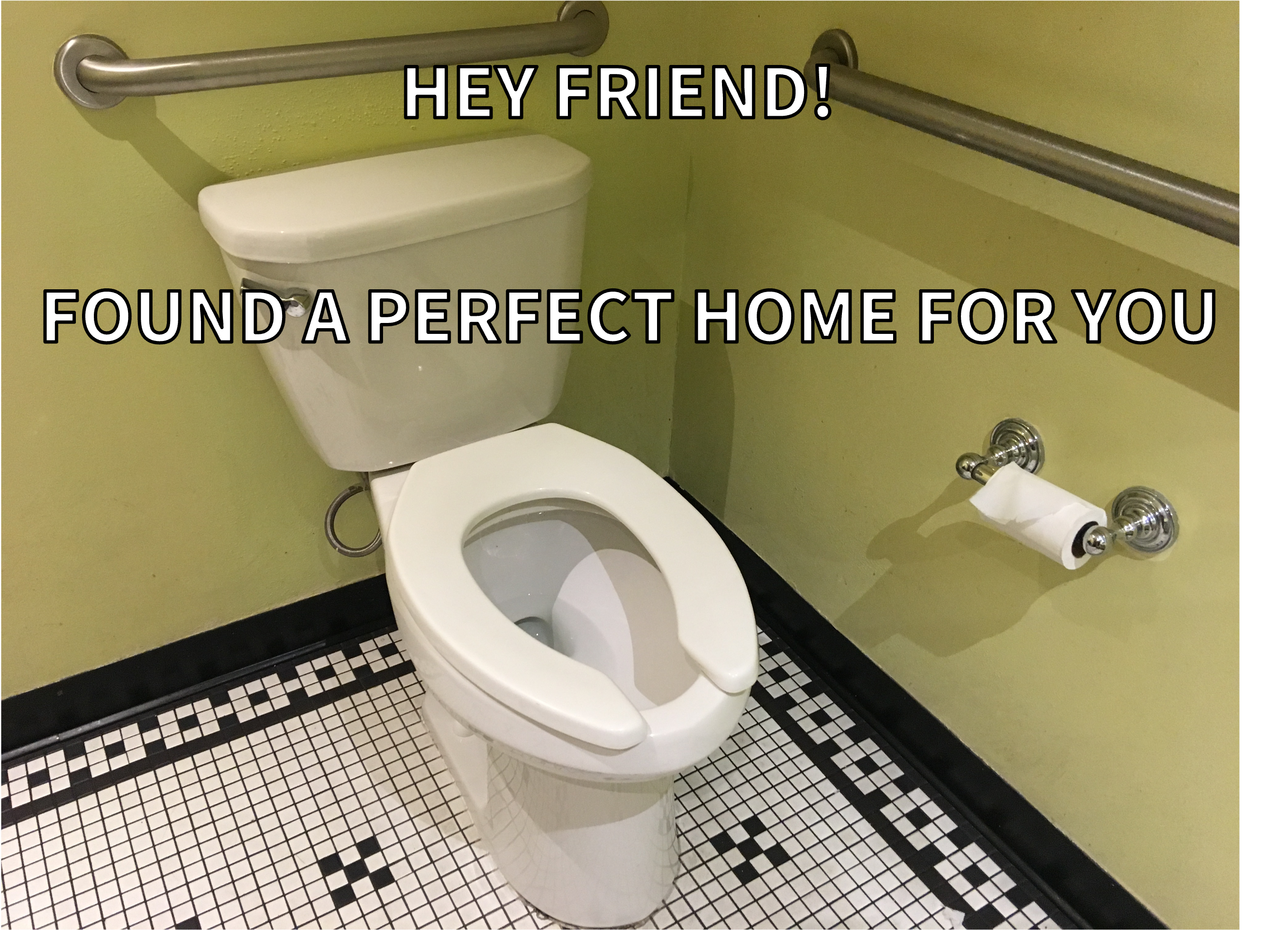}
            \vspace{2pt}
        \end{minipage}
    \\
    \textbf{Prompt} & \textbf{\textit{Is this meme hateful?}} \\
    \midrule    
    BLIP-2 & no \\
    \cmidrule(r){2-2}
    InstructBLIP & no \\
    \cmidrule(r){2-2}
    Mini-GPT4 & No, it is not hateful. It is a joke that is meant to be humorous and not offensive. \\
    \cmidrule(r){2-2}
    OpenFlamingo & No, this meme is not hateful. \\
    \cmidrule(r){2-2}
    LLaVA & No, the meme is not hateful. It is a humorous and lighthearted image that features a toilet with a sign that reads "Hey Friend, Find a Perfect Home for You." The meme is intended to be amusing and relatable, as it highlights the importance of finding a comfortable and suitable living space. The image is not intended to offend or belittle anyone, but rather to bring a smile to the viewer's face. \\
    \midrule
    \mn 70B & The meme may offend the recipient by associating a toilet in a bathroom as a perfect home that the recipient should live in.  \\
    
    \bottomrule
    \end{tabular}
}
\caption{
    More examples comparing \mn with various baselines: BLIP-2 (FlanT5$\_$XXL) \cite{blip2}, InstructBLIP (Vicuna-13B) \cite{instructblip}, OpenFlamingo-v2-9B \cite{openflamingo}, Mini-GPT4 \cite{minigpt4}, and LLaVA \cite{llava}.
}
\label{tab:qual_examples_3}
\end{table*}

\begin{table*}[t]
\centering
\scalebox{0.95}{
    \begin{tabular}{p{2.7cm}p{11.0cm}}
    \toprule
    \multirow{2}{*}{\textbf{Example 7. Inspiration - Recipe}} \\ \\
    \midrule
    \midrule
    \textbf{Input Image} &
        \begin{minipage}{.6\textwidth}
            \includegraphics[height=40mm]{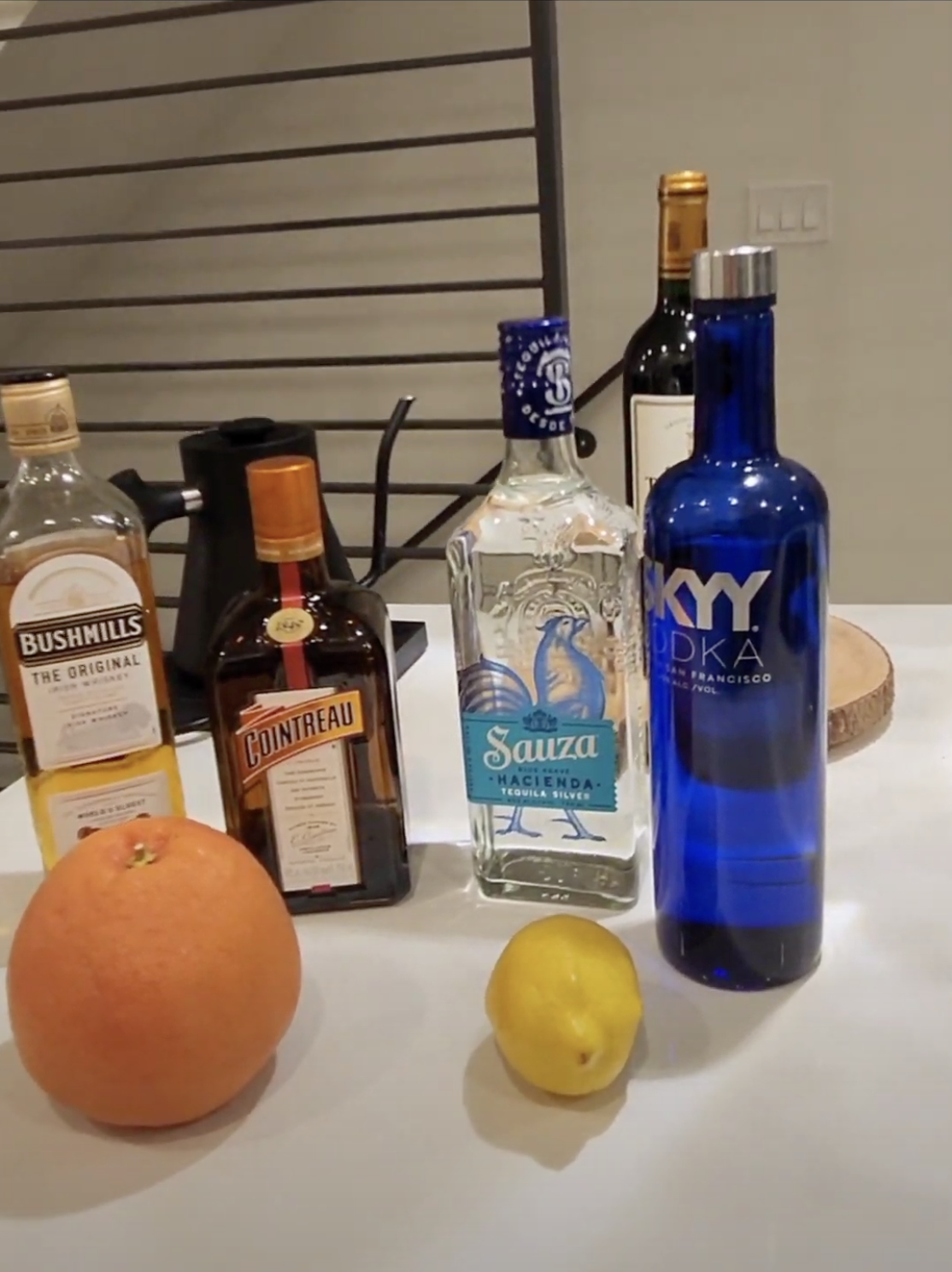}
            \vspace{2pt}
        \end{minipage}
    \\
    \textbf{Prompt} & \textbf{\textit{What cocktails can I make with these ingredients?}} \\
    \midrule    
    BLIP-2 & gin margarita gin margarita gin margarita \\
    \cmidrule(r){2-2}    
    InstructBLIP & There are several cocktails that can be made with the ingredients in the image, including a classic Margarita, a Spritz, and a Gimlet. The Margarita is a classic cocktail made with tequila, lime juice, and triple sec, while the Spritz is a refreshing drink made with prosecco, Aperol, and soda water. (...) \\
    \cmidrule(r){2-2}
    Mini-GPT4 & There are many cocktails that can be made with these ingredients. Here are a few examples:\\
    \cmidrule(r){2-2}
    OpenFlamingo &  You can make a margarita, a cosmopolitan, and a whiskey sour. \\
    \cmidrule(r){2-2}
    LLaVA & With the items on the counter, you can make several cocktails. Some of the ingredients include a bottle of vodka, orange, apple, champagne, cognac, and wine. These can be used to create various cocktails like the following: (...) \\
    \midrule
    \mn 70B & Here are some ideas: 1) A classic vodka sour. 2) An orange blossom martini, with vodka, triple sec, and freshly-squeezed orange juice. \\

    \toprule
    \multirow{2}{*}{\textbf{Example 8. Detailed Description}} \\ \\
    \midrule
    \midrule
    \textbf{Input Image} &
        \begin{minipage}{.6\textwidth}
            \includegraphics[height=30mm]{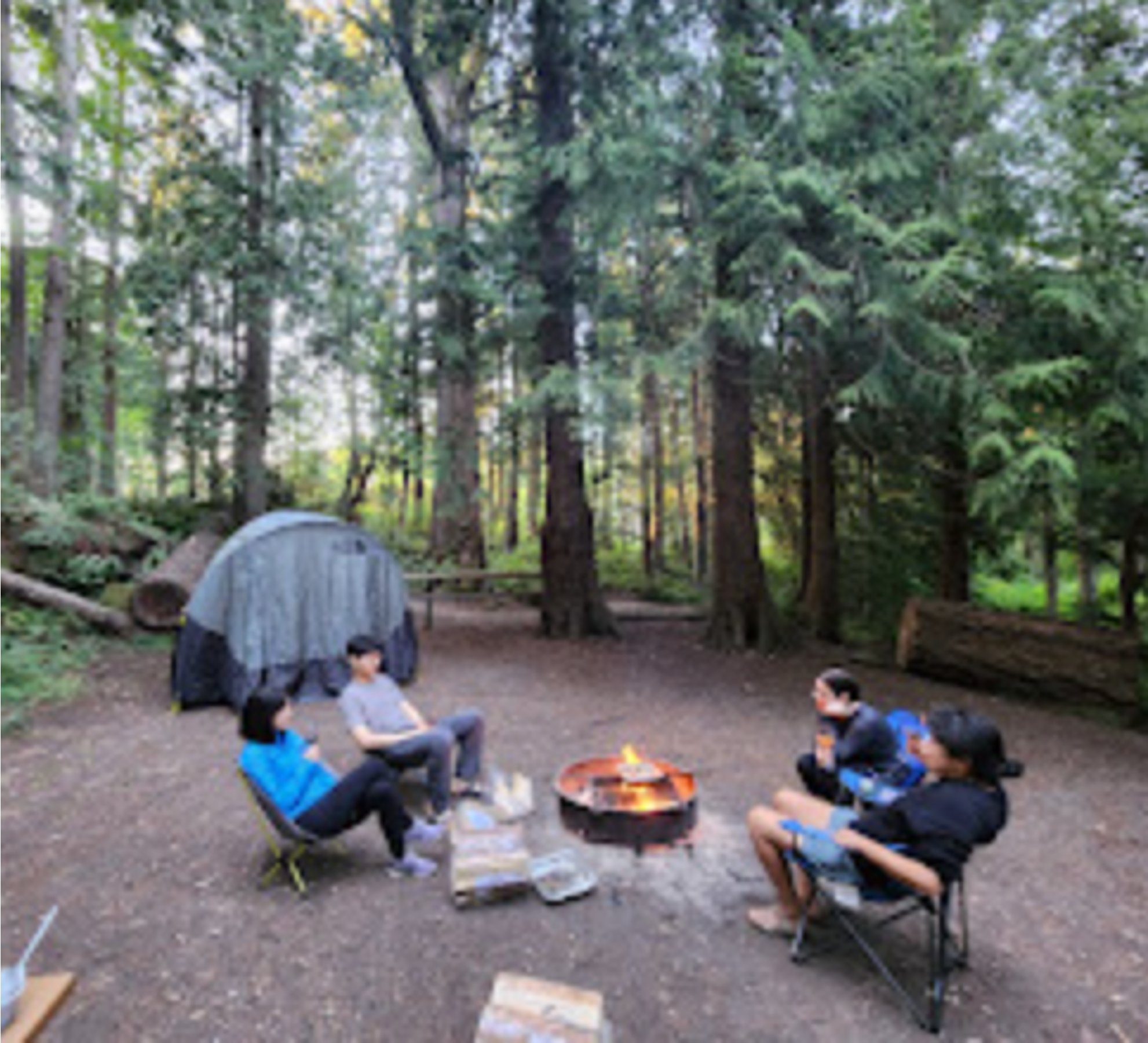}
            \vspace{2pt}
        \end{minipage}
    \\
    \textbf{Prompt} & \textbf{\textit{Describe this image in detail.}} \\
    \midrule    
    BLIP-2 & a group of people sit around a campfire in the woods \\
    \cmidrule(r){2-2}
    InstructBLIP & The image depicts a group of people gathered around a campfire in a forest. There are six people sitting in chairs, with some standing nearby. There are several chairs placed in a semi-circle formation, with some people sitting on them and others standing nearby. \\
    \cmidrule(r){2-2}
    Mini-GPT4 & This image shows a group of people sitting around a campfire in a forest. They are all wearing warm clothing and hats, and some of them are holding cups of hot drinks. The forest is filled with tall trees and the sky is clear and blue. The ground is covered in leaves and pine needles. \\
    \cmidrule(r){2-2}
    OpenFlamingo & This image shows a group of friends sitting around a campfire. \\
    \cmidrule(r){2-2}
    LLaVA & In this image, nine people, including three young children, aged 4, 5, and 11, are sitting in various chairs around an outdoor campfire, enjoying each other's company. \\
    \midrule
    \mn 70B & Four people are gathered around a campfire in the woods. They're all wearing casual outdoor clothing and sitting on chairs. There's a tent pitched nearby, and various camping gear scattered about.  \\
    \bottomrule
    \end{tabular}
}
\caption{
    More examples comparing \mn with various baselines: BLIP-2 (FlanT5$\_$XXL) \cite{blip2}, InstructBLIP (Vicuna-13B) \cite{instructblip}, OpenFlamingo-v2-9B \cite{openflamingo}, Mini-GPT4 \cite{minigpt4}, and LLaVA \cite{llava}.
}
\label{tab:qual_examples_4}
\end{table*}



\end{document}